\pdfoutput=1

\documentclass[11pt]{article}

\usepackage[]{acl}

\usepackage{times}
\usepackage{latexsym}

\usepackage[T1]{fontenc}

\usepackage[utf8]{inputenc}

\usepackage{microtype}

\usepackage{graphicx}
\usepackage{longtable}
\usepackage{multirow, makecell}
\usepackage{booktabs}
\usepackage{soul}
\usepackage{bbm}
\usepackage{hyperref}
\usepackage{amsmath}
\usepackage{amsthm}
\usepackage{amsfonts}
\usepackage{comment}
\usepackage{subcaption}
\usepackage{graphicx}

\usepackage{xpatch}
\makeatletter
\AtBeginDocument{\xpatchcmd{\@thm}{\thm@headpunct{.}}{\thm@headpunct{}}{}{}}
\makeatother

\theoremstyle{definition}

\newif\ifcomments
\commentstrue
\ifcomments
    \newcommand\sanjay[1]{{\color{blue}\{\textit{#1}\}$_{sanjay}$}}
    \newcommand\will[1]{{\color{purple}\{\textit{#1}\}$_\text{will}$}}
    \newcommand\trevor[1]{{\color{pink}\{\textit{#1}\}$_\text{trevor}$}}
    \newcommand\sameer[1]{{\color{magenta}\{\textit{#1}\}$_\text{sameer}$}}
    \newcommand\matt[1]{{\color{teal}\{\textit{#1}\}$_\text{matt}$}}
    \newcommand{\anja}[1]{\textcolor{red}{Anja:\ #1}}
    \newcommand{\medhini}[1]{\textcolor{red}{Anja:\ #1}}
\else
    \newcommand\sanjay[1]{}
    \newcommand\will[1]{}
    \newcommand\trevor[1]{}
    \newcommand\sameer[1]{}
    \newcommand\matt[1]{}
    \newcommand\anja[1]{}
    \newcommand\medhini[1]{}
\fi

\title{ReCLIP: A Strong Zero-Shot Baseline for \\Referring Expression Comprehension}

 \author{\makecell{Sanjay Subramanian\thanks{\hspace{0.3em} This work was done while Sanjay, Will, and Matt were affiliated with AI2.}$^{*1}$ ~~~~~~~ William Merrill$^{2}$ ~~~~~~~ Trevor Darrell$^{1}$ \\ Matt Gardner$^{3}$ ~~~~~ Sameer Singh$^{4,5}$ ~~~~~ Anna Rohrbach$^{1}$} \\ 
$^{1}$UC Berkeley\hspace{5mm}
$^{2}$New York University\hspace{5mm}
$^{3}$Microsoft Semantic Machines\hspace{5mm}
$^{4}$UC Irvine \\
$^{5}$Allen Institute for AI (AI2) \\
\texttt{\makecell{\{sanjayss,trevordarrell,anna.rohrbach\}@berkeley.edu, \\wcm9940@nyu.edu,
mattgardner@microsoft.com, sameer@uci.edu}}}

\begin{document}
\maketitle
\begin{abstract}
Training a referring expression comprehension (ReC) model for a new visual domain requires collecting referring expressions, and potentially corresponding bounding boxes, for images in the domain. While large-scale pre-trained models are useful for image classification across domains, it remains unclear if they can be applied in a zero-shot manner to more complex tasks like ReC.
We present ReCLIP, a simple but strong \emph{zero-shot} baseline that repurposes CLIP, a state-of-the-art large-scale model, for ReC.
Motivated by the close connection between ReC and CLIP's contrastive pre-training objective, the first component of ReCLIP is a region-scoring method that isolates object proposals via cropping and blurring, and passes them to CLIP. However, through controlled experiments on a synthetic dataset, we find that CLIP is largely incapable of performing spatial reasoning off-the-shelf.
Thus, the second component of ReCLIP is a spatial relation resolver that handles several types of spatial relations.
We  reduce the gap between zero-shot baselines from prior work and supervised  models by as much as 29\% on RefCOCOg, and on RefGTA (video game imagery), ReCLIP's relative improvement over supervised ReC models trained on real images is 8\%.
\end{abstract}

\section{Introduction}
\label{sec:intro}

\begin{figure}[t]
    \centering
    \begin{subfigure}{0.34\textwidth}
        \centering
        \includegraphics[width=0.9\textwidth]{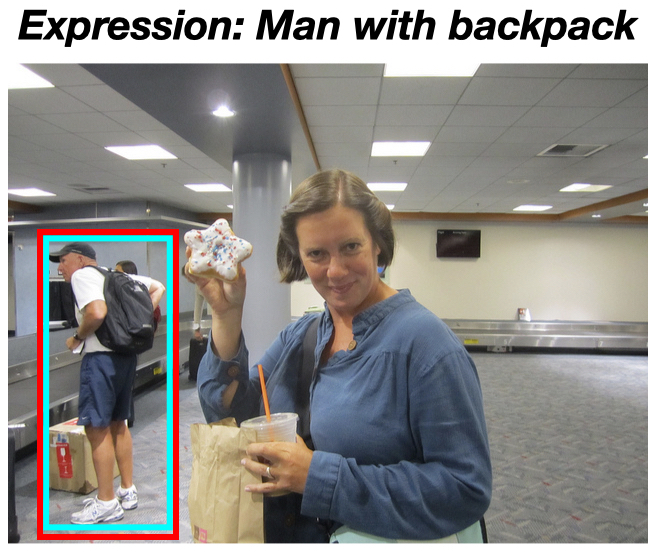}
        \vspace{-2mm}
        \caption{RefCOCO+~\cite{refcoco+}}
    \end{subfigure}
    \vspace{3mm}
    \begin{subfigure}{0.36\textwidth}
        \includegraphics[width=\textwidth]{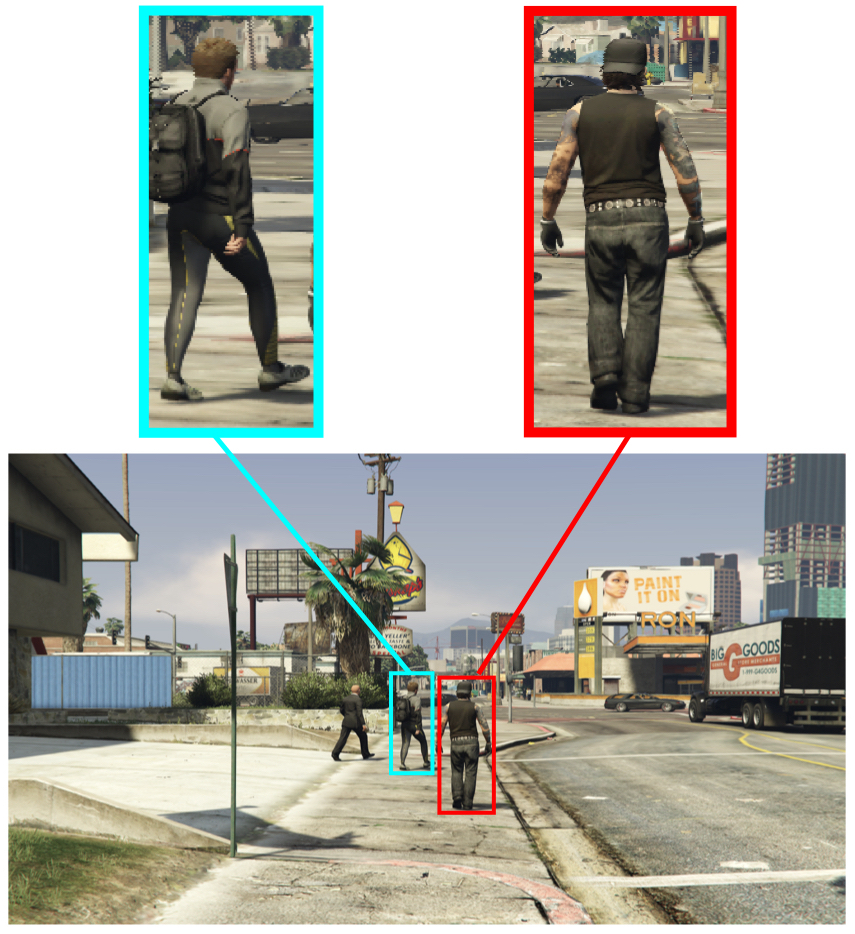}
        \vspace{-0.4cm}
        \caption{RefGTA~\cite{refgta}}
    \end{subfigure}
    \vspace{-0.5cm}
    \caption{Predictions from ReCLIP (\textcolor{cyan}{cyan}) and UNITER-Large~\citep{uniter} (\textcolor{red}{red}) for the same referring expression on images from two visual domains. UNITER-Large fails on the GTA (video game) domain, while ReCLIP selects the correct proposal in both cases. Close-ups of the two GTA boxes are shown.}
    \vspace{-0.4cm}
    \label{fig:teaser_fig1}
\end{figure}

Visual referring expression comprehension (ReC)---the task of localizing an object in an image given a textual referring expression---has applications in a broad range of visual domains. For example, ReC is useful for guiding a robot in the real world \citep{shridhar} and also for creating natural language interfaces for software applications with visuals \citep{Wichers2018ResolvingRE}. Though the task is the same across domains, the domain shift is problematic for supervised referring expression models, as shown in Figure~\ref{fig:teaser_fig1}: the same simple referring expression is localized correctly in the training domain but incorrectly in a new domain.

\begin{figure*}[t]
    \centering
    \includegraphics[width=\textwidth]{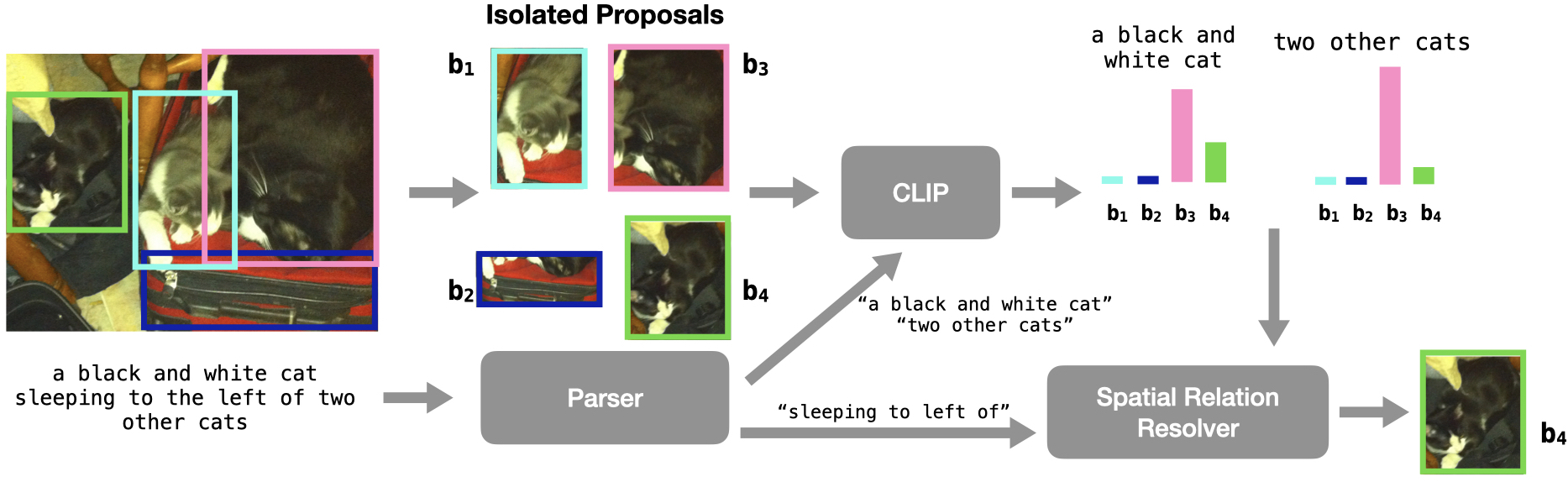}
    \vspace{-0.7cm}
    \caption{Overview of ReCLIP. Given object proposals, we isolate the corresponding image regions by cropping and blurring (only cropping shown here). Using a parser, we extract the noun chunks of the expression. For each noun chunk, CLIP outputs a distribution over proposals. The relations from the parser and CLIP's probabilities are combined by a spatial relation resolver to select the final proposal. In this example, CLIP ranks $b_3$ highest for both noun chunks, but using the relation resolver we obtain the correct answer $b_4$.}
    \vspace{-0.3cm}
    \label{fig:intro_fig1}
\end{figure*}

Collecting task-specific data in each domain of interest is expensive. Weakly supervised ReC \citep[]{grounder} 
partially addresses this issue, since it does not require the ground-truth box for each referring expression, but it still assumes the availability of referring expressions paired with images and trains on these. Given a large-scale pre-trained vision and language model and a method for doing ReC zero-shot---i.e. without any additional training---practitioners could save a great deal of time and effort. Moreover, as pre-trained models have become more accurate via scaling \citep{scaling}, fine-tuning the best models has become prohibitively expensive--and sometimes infeasible because the model is offered only via API, e.g. GPT-3 \citep{gpt3}.

Pre-trained vision and language models like CLIP \citep{clip} achieve strong zero-shot performance in image classification across visual domains \citep{align} and in object detection \citep{vild}, but the same success has not yet been achieved in tasks requiring reasoning over vision and language. For example, \citet{Shen2021HowMC} show that a straightforward zero-shot approach for VQA using CLIP performs poorly. Specific to ReC, \citet{cpt} introduce a zero-shot approach via Colorful Prompt Tuning (CPT), which colors object proposals and references the color in the text prompt to score proposals, but this has low accuracy. In both of these cases, the proposed zero-shot method is not aligned closely enough with the model's pre-training task of matching naturally occurring images and captions.

In this work, we propose ReCLIP, a simple but strong new baseline for zero-shot ReC. ReCLIP, illustrated in Figure~\ref{fig:intro_fig1}, has two key components: a method for scoring object proposals using CLIP and a method for handling spatial relations between objects. Our method for scoring region proposals, Isolated Proposal Scoring (IPS), effectively reduces ReC to the contrastive pre-training task used by CLIP and other models. Specifically, we propose to isolate individual proposals via cropping and blurring the images and to score these isolated proposals with the given expression using CLIP. 

To handle relations between objects, we first consider whether CLIP encodes the spatial information necessary to resolve these relations. We show through a controlled experiment on CLEVR images \citep{clevr} that CLIP and another pre-trained model ALBEF \citep{albef} are unable to perform its pre-training task on examples that require spatial reasoning. 

Thus, any method that solely relies on these models is unlikely to resolve spatial relations accurately. Consequently, we propose spatial heuristics for handling spatial relations in which an expression is decomposed into subqueries, CLIP is used to compute proposal probabilities for each subquery, and the outputs for all subqueries are combined with simple rules.

On the standard RefCOCO/g/+ datasets~\cite{refcocog,refcoco+}, we find that ReCLIP outperforms CPT \citep{cpt} by about 20\%. Compared to a stronger GradCAM \citep{gradcam} baseline, ReCLIP obtains better accuracy on average and has less variance across object types. Finally, in order to illustrate the practical value of zero-shot grounding, we also demonstrate that our zero-shot method surpasses the out-of-domain performance of state-of-the-art supervised ReC models. We evaluate on the RefGTA dataset \citep{refgta}, which contains images from a video game (out of domain for models trained only on real photos). Using ReCLIP and an object detector trained outside the target domain, we outperform UNITER-Large \citep{uniter} (using the same proposals) and MDETR \citep{mdetr} by an absolute 4.5\% (relative improvement of 8\%).

In summary, our contributions include: (1) ReCLIP, a zero-shot method for referring expression comprehension, (2) showing that CLIP has low zero-shot spatial reasoning performance, and (3) a comparison of our zero-shot ReC performance with the out-of-domain performance of state-of-the-art fully supervised ReC systems.\footnote{Our code is available at \url{https://www.github.com/allenai/reclip}.}
\section{Background}
\label{sec:methods}
In this section, we first describe the task at hand (\S\ref{sec:task}) and introduce CLIP, the pre-trained model we primarily use (\S\ref{sec:pretrained_models}). We then describe two existing methods for scoring region proposals using a pre-trained vision and language model: colorful prompt tuning (\S\ref{sec:cpt}) and GradCAM (\S\ref{sec:gradcam}).
\subsection{Task description}
\label{sec:task}
In referring expression comprehension (ReC), the model is given an image and a textual referring expression describing an entity in the image. The goal of the task is to select the object (bounding box) that best matches the expression. As in much of the prior work on REC, we assume access to a set of object proposals $b_1,b_2,...,b_n$, each of which is a bounding box in the image. Task accuracy is measured as the percentage of instances for which the model selects a proposal whose intersection-over-union (IoU) with the ground-truth box is at least 0.5. In this paper, we focus on the \emph{zero-shot} setting in which we apply a pre-trained model to ReC without using any training data for the task.

\subsection{Pre-trained model architecture}
\label{sec:pretrained_models}
The zero-shot approaches that we consider are general in that the only requirement for the pre-trained model is that when given a \emph{query} consisting of an image and text, it computes a score for the similarity between the image and text. In this paper, we primarily use CLIP \citep{clip}. We focus on CLIP because it was pre-trained on 400M image-caption pairs collected from the web\footnote{This dataset is not public.} and therefore achieves impressive zero-shot image classification performance on a variety of visual domains.
CLIP has an image-only encoder, which is either a ResNet-based architecture \citep{resnet} or a visual transformer \citep{vit}, and a text-only transformer. We mainly use the RN50x16 and ViT-B/32 versions of CLIP. The image encoder takes the raw image and produces an image representation $\mathbf{x} \in \mathbb{R}^d$, and the text transformer takes the sequence of text tokens and produces a text representation $\mathbf{y} \in \mathbb{R}^d$. In CLIP's contrastive pre-training task, given a batch of $ N $ images and matching captions, each image must be matched with the corresponding text. The model's probability of matching image $ i $ with caption $ j $ is given by $ \exp(\beta\mathbf{x_i}^T\mathbf{y_j})/\sum_{k=1}^N \exp(\beta\mathbf{x_i}^T\mathbf{y_k})$, where $\beta$ is a hyperparameter.\footnote{$\mathbf{x_i}$ and $\mathbf{y_i}$ are normalized before the dot product.}

We now describe two techniques from prior work for selecting a proposal using a pre-trained model.

\subsection{Colorful Prompt Tuning (CPT)}
\label{sec:cpt}
The first baseline from prior work that we consider is colorful prompt tuning (CPT), proposed by \citet{cpt} \footnote{CPT is the name given by \citet{cpt}, but note that we do not perform few-shot/supervised tuning.}: they shade proposals with different colors and use a masked language prompt in which the referring expression is followed by ``in [MASK] color''. The color with the highest probability from a pre-trained masked language model (MLM) (VinVL; \citep{vinvl}) is then chosen. In order to apply this method to models like CLIP, that provide image-text scores but do not offer an MLM, we create a version of the input image for each proposal, where the proposal is transparently shaded in red.\footnote{Specifically, we use the RGB values (240, 0, 30) and transparency 127/255 that \citet{cpt} say works best with their method. An example is shown in Appendix~\ref{app:cpt_example}.} Our template for the input text is ``[referring expression] is in red color.'' Since we have adapted CPT for non-MLM models, we refer to this method as \emph{CPT-adapted} in the experiments.

\subsection{Gradient-based visualizations}
\label{sec:gradcam}
The second baseline from prior work that we consider is based on gradient-based visualizations, which are a popular family of techniques for understanding, on a range of computer vision tasks, which part(s) of an input image are most important to a model's prediction. We focus on the most popular technique in this family, GradCAM \citep{gradcam}. Our usage of GradCAM follows \citet{albef}, in which GradCAM is used to perform weakly supervised referring expression comprehension using the ALBEF model, and \citet{chefer2021generic}. In our setting, for a given layer in a visual transformer, we take the layer's class-token (CLS) attention matrix $M \in \mathbb{R}^{h, w}$. The spatial dimensions $h$ and $w$ are dependent on the model's architecture and are generally smaller than the input dimensions of the image. Then the GradCAM is computed as $G = M \odot \frac{\partial L}{\partial M}$, where $L$ is the model's output logit (the similarity score for the image-text pair) and $\odot$ denotes elementwise multiplication. The procedure for applying GradCAM when the visual encoder is a convolutional network is similar\footnote{The convolutional version, following \citet{gradcam}, applies global average pooling to the gradients, unlike the transformer version.}; in place of the attention matrix, we use the activations of the final convolutional layer. Next, we perform a bicubic interpolation on $G$ so that it has the same dimensions as the input image. Finally, we compute for each proposal $b_i = (x_1, y_1, x_2, y_2)$ the score $\frac{1}{A^\alpha}\sum_{i=x_1}^{x_2}\sum_{j=y_1}^{y_2} G[i,j]$, where $A$ is the area of the image and $\alpha$ is a hyperparameter, and we choose the proposal with the highest score.
\section{ReCLIP}
ReCLIP consists of two main components: (1) a region-scoring method that is different from CPT and GradCAM and (2) a rule-based relation resolver. In this section, we first describe our region scoring method (\S\ref{sec:ips}). However, using controlled experiments on a synthetic dataset, we find that CLIP has poor zero-shot spatial reasoning performance (\S\ref{sec:clevr_exp}). Therefore, we propose a system that uses heuristics to resolve spatial relations (\S\ref{sec:relations}).

\subsection{Isolated Proposal Scoring (IPS)}
\label{sec:ips}
Our proposed method, which we call \emph{isolated proposal scoring}, is based on the observation that ReC is similar to the contrastive learning task with which models like CLIP are pre-trained, except that rather than selecting one out of several images to match with a given text, we must select one out of several image regions. Therefore, for each proposal, we create a new image in which that proposal is isolated. We consider two methods of isolation -- \emph{cropping} the image to contain only the proposal and \emph{blurring} everything in the image except for the proposal region. For blurring, we apply a Gaussian filter with standard deviation $\sigma$ to the image RGB values. Appendix~\ref{app:ips_example} provides an example of isolation by blurring. The score for an isolated proposal is obtained by passing it and the expression through the pre-trained model. To use cropping and blurring in tandem, we obtain a score $s_{crop}$ and $s_{blur}$ for each proposal and use $s_{crop}+s_{blur}$ as the final score. This can be viewed as an ensemble of ``visual prompts,'' analogous to \citet{clip}'s ensembling of text prompts.

\begin{table}[t]
\centering
\captionsetup{font=footnotesize}
\resizebox{1.\linewidth}{!}
{
\begin{tabular}{lcccc}
\toprule
\multirow{2}[1]{*}{Model} & \multicolumn{1}{c}{Text-pair} & \multicolumn{1}{c}{Text-pair} & \multicolumn{1}{c}{Image-pair} & \multicolumn{1}{c}{Image-pair} \\
& \multicolumn{1}{c}{Spatial} & \multicolumn{1}{c}{Non-spatial} & \multicolumn{1}{c}{Spatial} & \multicolumn{1}{c}{Non-spatial} \\
\midrule
CLIP RN50x4 & 43.39 & 89.83 & 48.90 & 97.36 \\
CLIP RN50x16 & 51.19 & 89.83 & 50.22 & 96.48 \\
CLIP RN50x64 & 47.80 & 94.58 & 51.54 & 97.36 \\
CLIP ViT-B/32 & 48.47 & 95.25 & 48.90 & 96.48 \\
CLIP ViT-B/16 & 50.51 & 92.54 & 50.22 & 96.92 \\
CLIP ViT-L/14 & 52.88 & 96.27 & 50.66 & 94.27 \\
\bottomrule
\end{tabular}
}
\caption{Accuracy on CLEVR image-text matching task. CLIP performs well on the non-spatial version of the task but poorly on the spatial version. Text-pair tasks have 295 instances each; image-pair tasks have 227 instances each.}
\label{tab:clevr}
\end{table}

\subsection{Can we use CLIP to resolve spatial relations?}
\label{sec:clevr_exp}
A key limitation in Isolated Proposal Scoring is that relations between objects in different proposals are not taken into account. For example, in Figure ~\ref{fig:intro_fig1}, the information about the spatial relationships among the cats is lost when the proposals are isolated. In order to use CLIP to decide which object has a specified relation to another object, the model's output must encode the spatial relation in question.
Therefore, we design an experiment to determine whether a pre-trained model, such as CLIP, can understand spatial relations within the context of its pre-training task. We generate synthetic images using the process described for the CLEVR dataset \citep{clevr}. These scenes include three shapes--spheres, cubes, and cylinders--and eight colors--gray, blue, green, cyan, yellow, purple, brown, red.

In the \emph{text-pair} version of our tasks, using the object attribute and position information associated with each image, we randomly select one of the pairwise relationships between objects--left, right, front, or behind--and construct a sentence fragment based on it. For example: ``A blue sphere to the left of a red cylinder.'' We also write a distractor fragment that replaces the relation with its opposite. In this case, the distractor would be ``A blue sphere to the right of a red cylinder.'' The task, similar to the contrastive and image-text matching tasks used to pre-train these models, is to choose the correct sentence given the image. As a reference point, we also evaluate on a control (non-spatial) task in which the correct text is a list of the scene's objects and the distractor text is identical except that one object is swapped with a random object not in the scene. For example, if the correct text is ``A blue sphere and a red cylinder,'' then the distractor text could be ``A blue sphere and a blue cylinder.''

In the \emph{image-pair} version of our tasks, we have a single sentence fragment constructed as described above for the spatial and control (non-spatial) tasks and two images such that only one matches the text. Appendix~\ref{app:synthetic} shows examples of these tasks.

CLIP's performance on these tasks is shown in Table~\ref{tab:clevr}. Similar results for the pre-trained model ALBEF \citep{albef} are shown in Appendix~\ref{app:albef_relations} While performance on the control task is quite good, accuracy on the spatial task is not so different from random chance (50\%). This indicates that the model scores of image-text pairs largely do not take spatial relations into account. 

\subsection{Spatial Relation Resolver}
\label{sec:relations}
Since CLIP lacks sensitivity to spatial relations, we propose to decompose complex expressions into simpler primitives. The basic primitive is a predicate applying to an object, which we use CLIP to answer. The second primitive is a spatial relation between objects, for which we use heuristic rules.

\paragraph{Predicates} A predicate is a textual property that the referent must satisfy. For example, ``the cat'' and ``blue airplane'' are predicates. We write $P(i)$ to say that object $i$ satisfies the predicate $P$. We model $P$ as a categorical distribution over objects, and estimate $p(i) = \Pr[P(i)]$ with the pre-trained model using isolated proposal scoring (\S~\ref{sec:ips}).

\paragraph{Relations} We have already discussed the importance of binary spatial relations like ``the cat to the \emph{left} of the dog'' for the ReC task. We consider seven spatial relations--\emph{left}, \emph{right}, \emph{above}, \emph{below}, \emph{bigger}, \emph{smaller}, and \emph{inside}. We write $R(i, j)$ to mean that the relation $R$ holds between objects $i$ and $j$, and we use heuristics to determine the probability $r(i, j) = \Pr[R(i, j)]$. For example, for \emph{left}, we set $r(i, j)=1$ if the center point of box $i$ is to the left of the center point of box $j$ and $r(i,j)=0$ otherwise. \S\ref{app:heuristics} describes all relation semantics.

\paragraph{Superlative Relations} We also consider superlatives, which refer to an object that has some relation to all other objects satisfying the same predicate, e.g. ``leftmost dog''.
We handle superlatives as a special case of relations where the empty second argument is filled by copying the predicate specifying the first argument.
Thus, ``leftmost dog'' effectively finds the dog that is most likely to the left of other dog(s).
Our set of superlative relation types is the same as our set of relation types, excluding \emph{inside}.

\begin{figure}
    \centering
    \captionsetup{font=footnotesize}
    \includegraphics[width=\columnwidth]{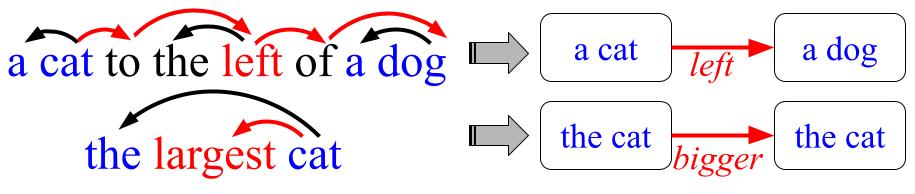}
    \caption{Example extraction of semantic trees from dependency parses. Predicate text in blue. Red arcs show paths contributing spatial relation \emph{left} and superlative \emph{largest}. For the superlative, we create a parent node with the original node as the only child, effectively converting it into a relation.}
    \label{fig:entity-tree}
\end{figure}

\paragraph{Semantic Trees} Having outlined the semantic formalism underlying our method, we can describe it procedurally. We first use spaCy \citep{spacy} to build a dependency parse for the expression. As illustrated in Figure~\ref{fig:entity-tree}, we extract a semantic tree from the dependency parse, where each noun chunk becomes a node, and dependency paths between the heads of noun chunks become relations between entities based on the keywords they contain. See \S\ref{app:extract} for extraction details. In cases where none of our relation/superlative keywords occur in the text, we simply revert to the plain isolated proposal scoring method using the full text.

In the tree, each node $N$ contains a predicate $P_N$ and has a set of children; an edge $(N, N')$ between $N$ and its child $N'$ corresponds to a relation $R_{N,N'}$.
For example, as shown in \autoref{fig:entity-tree}, ``a cat to the left of a dog'' would be parsed as a node containing the predicate ``a cat'' connected by the relation \emph{left} to its child corresponding to ``a dog''. We define $\pi_N(i)$ as the probability that node $N$ refers to object $i$, and compute it recursively. For each node $N$, we first set $\pi_N(i) = p_N(i)$ and then iterate through each child $N'$ and update $\pi_N(i)$ as follows\footnote{Superlatives of a node are processed after all its relations.}:
\vspace{-0.2cm}
\begin{align*}
    \pi_N'(i) &\propto \pi_N(i) \sum_{j}\mathrm{Pr} \left[R_{N,N'}(i, j) \wedge P_{N'}(j) \right] \\
    &\propto \pi_N(i) \sum_{j} r_{N,N'}(i,j)\pi_{N'}(j) .
\end{align*}
The last line makes the simplifying assumption that all predicates and relations are independent.\footnote{We write $\propto$ because $\pi'_N(i)$ is normalized to sum to $1$.}

To compute our final output, we ensemble the distribution $\pi_{root}$ for the root node with the output of plain isolated proposal scoring (with the whole input expression) by multiplying the proposal probabilities elementwise. This method gives us a principled way to combine predicates ($P_N$) with spatial relational constraints ($R_{N,N'}$) for each node $N$.
\section{Experiments}

\label{sec:experiments}
\begin{table*}[t]
\small
\centering
\captionsetup{font=footnotesize}
\resizebox{1.0\textwidth}{!}{
\begin{tabular}{lcc|ccc|ccc}
\toprule
\multirow{2}[3]{*}{{\bf Model}} & \multicolumn{2}{c}{RefCOCOg} & \multicolumn{3}{c}{RefCOCO+} & \multicolumn{3}{c}{RefCOCO} \\
& \multicolumn{1}{c}{\textbf{Val}} &  \multicolumn{1}{c}{\textbf{Test}} &  \multicolumn{1}{c}{\textbf{Val}} &  \multicolumn{1}{c}{\textbf{TestA}} &  \multicolumn{1}{c}{\textbf{TestB}}  & \multicolumn{1}{c}{\textbf{Val}} &  \multicolumn{1}{c}{\textbf{TestA}} &  \multicolumn{1}{c}{\textbf{TestB}}\\
\midrule
Random & 18.12 & 19.10 & 16.29 & 13.57 & 19.60 & 15.73 & 13.51 & 19.20 \\
\midrule
Supervised SOTA & 83.35 & 81.64 & 81.13 & 85.52 & 72.96 & 87.51 & 90.40 & 82.67 \\
\midrule
CPT-Blk w/ VinVL \citep{cpt} & 32.1 & 32.3 & 25.4 & 25.0 & 27.0 & 26.9 & 27.5 & 27.4 \\
CPT-Seg w/ VinVL \citep{cpt} & 36.7 & 36.5 & 31.9 & 35.2 & 28.8 & 32.2 & 36.1 & 30.3 \\
\midrule
\textbf{CLIP} & & & & & & &\\
CPT-adapted & 22.32 & 23.65 & 23.85 & 21.55 & 25.92 & 23.16 & 21.44 & 26.95 \\
GradCAM & 50.86 & 49.70 & 47.83 & \textbf{56.92} & 37.70 & 42.85 & \textbf{51.07} & 35.21 \\
ReCLIP w/o relations & 57.70 & 57.19 & 47.43 & 50.02 & 43.85 & 41.97 & 43.42 & 39.02 \\
ReCLIP & \textbf{59.33} & \textbf{59.01} & \textbf{47.87} & 50.10 & \textbf{45.10} & \textbf{45.78} & 46.10 & \textbf{47.07} \\
\midrule
\textbf{CLIP w/ Object Size Prior} & & & & & & & & \\
CPT-adapted & 28.98 & 30.14 & 26.64 & 25.13 & 27.27 & 26.08 & 25.38 & 28.03 \\
GradCAM & 52.29 & 51.28 & 49.41 & 59.66 & 38.62 & 44.65 & 53.49 & 36.19 \\
ReCLIP w/o relations & 59.19 & 59.01 & 54.66 & 60.27 & 46.33 & 48.53 & 53.60 & 40.84 \\
ReCLIP & \underline{60.85} & \underline{61.05} & \underline{55.07} & \underline{60.47} & \underline{47.41} & \underline{54.04} & \underline{58.60} & \underline{49.54} \\
\bottomrule
\end{tabular}
}
\caption{Accuracy on the RefCOCOg, RefCOCO+ and RefCOCO datasets. ReCLIP outperforms other zero-shot methods on RefCOCOg. On RefCOCO+ and RefCOCO, ReCLIP is on par with or better than GradCAM on average and has lower variance between TestA and TestB, which correspond to different kinds of objects. When taking into account a prior on object size (filtering out objects smaller than 5\% of the image), GradCAM's advantage on the TestA splits is erased. Best zero-shot results in each column are in \textbf{bold}, and best zero-shot results using the size prior are \underline{underlined}. CLIP results use an ensemble of the RN50x16 and ViT-B/32 CLIP models. CPT-adapted is an adapted version of CPT-Blk. Supervised SOTA refers to MDETR \citep{mdetr}; we use the EfficientNet-B3 version. All methods except MDETR use detected proposals from MAttNet \citep{mattnet}. CPT-Seg uses Mask-RCNN segmentation masks from \citet{mattnet}.}
\label{tab:coco-results}
\end{table*}

\subsection{Datasets}
We compare ReCLIP to other zero-shot methods on \textbf{RefCOCOg} \citep{refcocog}, \textbf{RefCOCO} and \textbf{RefCOCO+} \citep{refcoco+}. These datasets use images from MS~COCO \citep{coco}. RefCOCO and RefCOCO+ were created in a two-player game, and RefCOCO+ is designed to avoid spatial relations. RefCOCOg includes spatial relations and has longer expressions on average. For comparing zero-shot methods with the out-of-domain performance of models trained on COCO, we use \textbf{RefGTA} \citep{refgta}, which contains images from the Grand Theft Auto video game. All referring expressions in RefGTA correspond to people, and the objects (i.e. people) tend to be much smaller on average than those in RefCOCO/g/+. 

\subsection{Implementation Details}
We use an ensemble of the CLIP RN50x16 and ViT-B/32 models (results for individual models are shown in Appendix~\ref{app:results}). We ensemble model outputs by adding together the logits from the two models elementwise before taking the softmax. GradCAM's hyperparameter $\alpha$ controls the effect of the proposal's area on its score. We select $\alpha=0.5$ for all models based on tuning on the RefCOCOg validation set. We emphasize that the optimal value of $\alpha$ for a dataset depends on the size distribution of ground-truth objects. ReCLIP also has a hyperparameter, namely the standard deviation $\sigma$. We try a few values on the RefCOCOg validation set and choose $\sigma=100$, as we show in Appendix~\ref{app:hyperparams}, isolated proposal scoring has little sensitivity to $\sigma$. As discussed by \citep{Perez2021TrueFL}, zero-shot experiments often use labeled data for model selection. Over the course of this work, we primarily experimented with the RefCOCOg validation set and to a lesser extent with the RefCOCO+ validation set. For isolated proposal scoring, the main variants explored are documented in our ablation study (\S\ref{sec:ablations}). Other techniques that we tried, including for relation-handling, and further implementation details are given in Appendix~\ref{app:implementation}.

\subsection{Results on RefCOCO/g/+}
\label{sec:coco_results}
Table~\ref{tab:coco-results} shows results on RefCOCO, RefCOCO+, and RefCOCOg. ReCLIP is better than the other zero-shot methods on RefCOCOg and RefCOCO and on par with GradCAM on RefCOCO+. However, GradCAM has a much higher variance in its accuracy between the TestA and TestB splits of RefCOCO+ and RefCOCO. We note that GradCAM's hyperparameter $\alpha$, controlling the effect of proposal size, was tuned on the RefCOCOg validation set, and RefCOCOg was designed such that boxes of referents are at least 5\% of the image area \citep{refcocog}. In the bottom portion of Table~\ref{tab:coco-results}, we show that when this 5\% threshold, a prior on object size for this domain, is used to filter proposals for both GradCAM and ReCLIP
, ReCLIP performs on par with/better than GradCAM on TestA. ReCLIP's spatial relation resolver helps on RefCOCOg and RefCOCO but not on RefCOCO+, which is designed to avoid spatial relations.

\subsection{Results on RefGTA}
Next, we evaluate on RefGTA to compare our method's performance to the out-of-domain accuracy of two state-of-the-art fully supervised ReC models: UNITER-Large \citep{uniter} and MDETR \citep{mdetr}.

Like ReCLIP, UNITER takes proposals as input.\footnote{UNITER requires features from the bottom-up top-down attention model \citep{butd}. We use \url{https://github.com/airsplay/py-bottom-up-attention} to compute the features for RefGTA.
We trained UNITER models on RefCOCO+ and RefCOCOg using features computed from this repository. On the RefCOCO+ validation set, the resulting model has an accuracy roughly 0.4\% less than that of a model trained and evaluated using the original features (when using ground-truth proposals).} We show results using ground-truth proposals and detections from UniDet \citep{unidet}, which is trained on the COCO, Objects365 \citep{objects365}, OpenImages \citep{openimages}, and Mapillary \citep{Mapillary} datasets. Following the suggestion of the UniDet authors, we use the confidence threshold of 0.5. MDETR does not take proposals as input.

Table~\ref{tab:refgta} shows our results. For methods that take proposals (all methods except MDETR), we consider two evaluation settings using UniDet--\emph{DT-P}, in which the detected proposals are filtered to have only proposals whose predicted class label is ``person'', and \emph{DT}, in which all detected proposals are considered. ReCLIP's accuracy is more than 15\% higher than the accuracy of UNITER-Large and roughly 5\% more than that of MDETR. ReCLIP also outperforms GradCAM by about 20\%, and the gap is larger when all UniDet proposals are considered. ReCLIP w/o relations is 1-2\% better than ReCLIP in the settings with ground-truth proposals and filtered UniDet proposals. One possible reason for this gap is that the objects of relations in the expressions could be non-people entities. When considering all UniDet proposals, the relation resolver in ReCLIP does not hurt accuracy much but also does not improve accuracy significantly--an additional challenge in this setting is that the number of proposals is dramatically higher. Appendix~\ref{app:qualitative_examples} shows qualitative examples of predictions on RefGTA.

\begin{table}[t]
\captionsetup{font=footnotesize}
\resizebox{1.\linewidth}{!}
{
\begin{tabular}{lccccccc}
\toprule
\multirow{2}[3]{*}{{\bf Model}} & \multicolumn{3}{c}{\textbf{Val}} & \multicolumn{3}{c}{\textbf{Test}} \\
& \multicolumn{1}{c}{GT} &  \multicolumn{1}{c}{DT-P} &  \multicolumn{1}{c}{DT} &  \multicolumn{1}{c}{GT} &  \multicolumn{1}{c}{DT-P}  & \multicolumn{1}{c}{DT} \\
\midrule
Random & 27.03 & 21.53 & 4.86 & 27.60 & 21.75 & 5.13 \\
UNITER-Large & & & & & & \\
\hspace{3mm}\emph{RefCOCO+} & 49.57 & 47.52 & 35.04 & 50.60 & 48.30 & 34.40 \\
\hspace{3mm}\emph{RefCOCOg} & 49.81 & 48.59 & 27.58 & 51.05 & 49.78 & 28.31 \\
MDETR & & & & & & \\
\hspace{3mm}\emph{RefCOCO+} & -- & -- & 38.49 & -- & -- & 39.02 \\
\hspace{3mm}\emph{RefCOCOg} & -- & -- & 38.29 & -- & -- & 39.13 \\
\hspace{3mm}\emph{Pretrained} & -- & -- & 54.91 & -- & -- & 56.60 \\
CLIP GradCAM & 51.90 & 51.03 & 33.66 & 51.53 & 50.73 & 34.51 \\
ReCLIP & 69.84 & 68.42 & 60.93 & 70.79 & 69.05 & \textbf{61.38} \\
\hspace{3mm}\emph{w/o relations} & \textbf{71.66} & \textbf{70.27} & \textbf{60.98} & \textbf{72.56} & \textbf{70.84} & 61.31 \\
\bottomrule
\end{tabular}
}
\captionof{table}{Accuracy on RefGTA dataset. ReCLIP w/o relations outperforms all other methods. \emph{GT} denotes use of ground-truth proposals; \emph{DT} denotes use of detected proposals; \emph{DT-P} denotes detected proposals filtered to have only people. Subscripts \emph{RefCOCO+}/\emph{RefCOCOg} indicate finetuning dataset; \emph{Pretrained} indicates a model that is not finetuned. MDETR does not take proposals as input, so the \emph{GT} and \emph{DT-P} columns are blank. We use the EfficientNet-B3 versions of MDETR. \textbf{Bold} indicates best score in a column.}
\vspace{-0.3cm}
\label{tab:refgta}
\end{table}

\subsection{Using another Pre-trained Model}
In order to determine how isolated proposal scoring (IPS) compares to GradCAM and CPT on other pre-trained models, we present results using ALBEF \citep{albef}. ALBEF offers two methods for scoring image-text pairs--the output used for its image-text contrastive (ITC) loss and the output used for its image-text matching (ITM) loss. The architecture providing the ITC output is very similar to CLIP--has only a shallow interaction between the image and text modalities. The ITM output is given by an encoder that has deeper interactions between image and text and operates on top of the ITC encoders' output. Appendix~\ref{app:albef_details} provides more details. The results, shown in Table~\ref{tab:albef_results}, show that with the ITC output, IPS performs better than GradCAM, but with the ITM output, GradCAM performs better. This suggests that IPS works well across models like CLIP and ALBEF ITC (i.e. contrastively pre-trained with shallow modality interactions) but that GradCAM may be better for models with deeper interactions.
\begin{table}[t]
\small
\centering
\captionsetup{font=footnotesize}
\resizebox{\linewidth}{!}
{
\begin{tabular}{lcccc}
\toprule
{{\bf Model}} & RefCOCOg & RefCOCO+(A) & RefCOCO+(B) \\
\midrule
\multicolumn{3}{l}{\textbf{ALBEF ITM} (Deep modality interaction)} \\
CPT-adapted & 24.99 & 26.83 & 26.43 \\
GradCAM & \textbf{55.92} & \textbf{61.75} & \textbf{42.79} \\
IPS & 55.21 & 51.82 &42.63 \\
\midrule
\multicolumn{3}{l}{\textbf{ALBEF ITC} (Shallow modality interaction)} \\
CPT-adapted & 21.10 & 19.00 & 21.33 \\
GradCAM & 47.53 & 44.60 & 36.00 \\
IPS & \textbf{54.07} & \textbf{45.90}& \textbf{39.58} \\
\bottomrule
\end{tabular}
}
\captionof{table}{Accuracy on RefCOCOg and RefCOCO+ test sets using ALBEF pre-trained model. IPS does best when using ALBEF's ITC architecture, while GradCAM is better for ITM.}
\label{tab:albef_results}
\end{table}

\subsection{Analysis}
\label{sec:ablations}
\begin{figure}[t]
    \centering
    \begin{subfigure}{0.5\textwidth}
        \centering
        \caption{ReCLIP is correct, while GradCAM is incorrect}
        \includegraphics[width=0.6\textwidth]{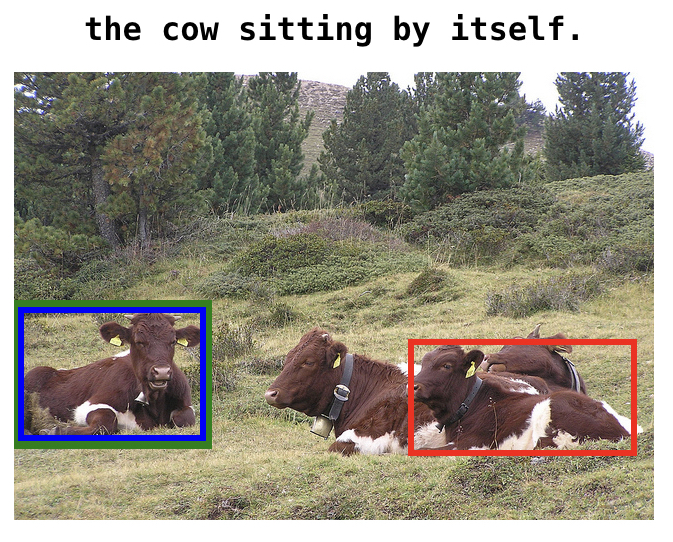}
        \label{fig:subfig_analysis1}
    \end{subfigure}
    \begin{subfigure}{0.5\textwidth}
        \centering
        \caption{Both ReCLIP and GradCAM are incorrect}
        \includegraphics[width=0.6\textwidth]{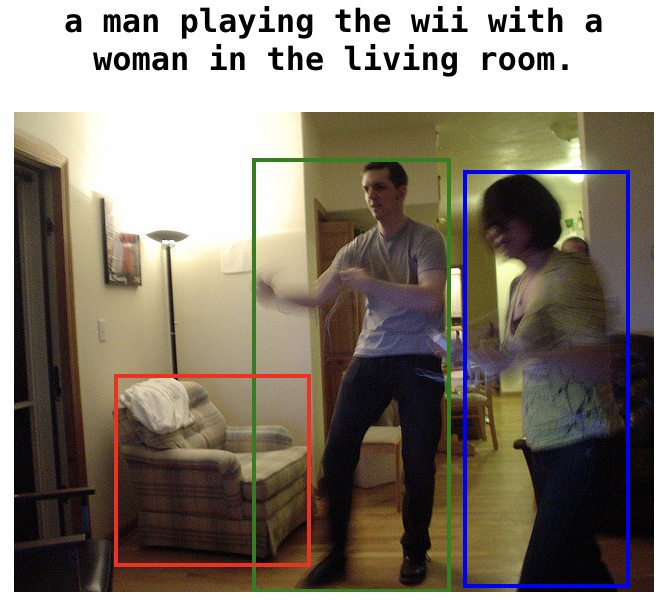}
        \label{fig:subfig_analysis2}
    \end{subfigure}
    \caption{RefCOCOg validation examples using ground-truth proposals. Ground-truth referents are \textcolor{green!45!black}{green}, ReCLIP predictions are \textcolor{blue}{blue}, and GradCAM predictions are \textcolor{red}{red}. In \ref{fig:subfig_analysis1}, ReCLIP makes the correct prediction based on local context. In \ref{fig:subfig_analysis2}, ReCLIP grounds an incorrect noun chunk from the expression.}
    \vspace{-0.4cm}
    \label{fig:analysis_fig1}
\end{figure}

\paragraph{Performance of IPS} Our results show that among the region scoring methods that we consider, IPS achieves the highest accuracy for contrastively pre-trained models like CLIP.
Figure~\ref{fig:subfig_analysis1} gives intuition for this---aside from an object's attributes, many referring expressions describe the local context around an object, and IPS focuses on this local context (as well as object attributes).

\begin{table}[t]
\small
\centering
\captionsetup{font=footnotesize}
\begin{tabular}{lcc}
\toprule
{\bf Isolation type} & RefCOCOg & RefCOCO+ \\
\midrule
Crop & 54.43 & 41.28 \\
Blur & 55.96 & 47.23 \\
$\max($Crop,Blur$)$ & 55.76 & 44.55 \\
Crop+Blur & \textbf{57.70} & \textbf{47.43} \\
\bottomrule
\end{tabular}
\captionof{table}{Ablation study of isolation types used to score proposals on Val splits of RefCOCOg/RefCOCO+, using detections from MAttNet \citep{mattnet}. Crop+Blur is best overall.}
\vspace{-0.4cm}
\label{tab:ablation}
\end{table}

Table~\ref{tab:ablation} shows that using both cropping and blurring obtains greater accuracy than either alone.

\paragraph{Error Analysis and Limitations} Although ReCLIP outperforms the baselines that we consider, there is a considerable gap between it and supervised methods. The principal challenge in improving the system is making relation-handling more flexible. There are several object relation types that our spatial relation resolver cannot handle; for instance, those that involve counting: ``the second dog from the right.'' Another challenge is in determining which relations require looking at multiple proposals. For instance, ReCLIP selects a proposal corresponding to the incorrect noun chunk in Figure~\ref{fig:subfig_analysis2} because the relation resolver has no rule for splitting an expression on the relation ``with.'' Depending on the context, relations like ``with'' may or may not require looking at multiple proposals, so handling them is challenging for a rule-based system.

In the RefCOCO+ validation set, when using detected proposals, there are 75 instances for which ReCLIP answers incorrectly but ReCLIP w/o relations answers correctly. We categorize these instances based on their likely sources of error: 4 instances are ambiguous (multiple valid proposals), in 7 instances the parser misses the head noun chunk, in 14 instances our processing of the parse leads to omissions of text when doing isolated proposal scoring (e.g. in ``girl sitting in back,'' the only noun chunk is ``girl,'' so this is the only text used during isolated proposal scoring), 52 cases in which there is an error in the execution of the heuristic (e.g. our spatial definition of a relation does not match the relation in the instance). (There are 2 instances for which we mark 2 categories.) The final category (``execution'') includes several kinds of errors, some examples of which are shown in Appendix~\ref{app:qualitative_examples}.

\section{Related Work}
\label{sec:related}

\paragraph{Referring expression comprehension} Datasets for ReC span several visual domains, including photos of everyday scenes \citep{refcocog,Kazemzadeh2014ReferItGameRT}, video games \citep{refgta}, objects in robotic context \citep{shridhar,Wang2021OCIDRefA3}, and webpages \citep{Wichers2018ResolvingRE}.

Spatial heuristics have been used in previous work \citep{moratz2006spatial}. Our work is also related to \citet{krishnamurthy-kollar-2013-jointly}, which similarly decomposes the reasoning process into a parsing step and visual execution steps, but the visual execution is driven by learned binary classifiers for each predicate type. In the supervised setting, prior work shows that using an external parser, as we do, leads to lower accuracy than training a language module jointly with the remainder of the model \citep{hu2017modeling}.

There is a long line of work in weakly supervised ReC, where at training time, pairs of referring expressions and images are available but the ground-truth bounding boxes for each expression are not \citep{grounder,arn,vc,ccl,Sun2021DiscriminativeTM}.
Our setting differs from the weakly supervised setting in that the model is not trained at all on the ReC task. \citet{zsg} discuss a zero-shot setting different from ours in which novel objects are seen at test time, but the visual domain stays the same.

\paragraph{Pre-trained vision and language models} Early pre-trained vision and language models \citep{lxmert,vilbert,uniter} used a cross-modal transformer \citep{transformer} and pre-training tasks like masked language modeling, image-text matching, and image feature regression. By contrast, CLIP and similar models \citep{clip,align} use a separate image and text transformer and a contrastive pre-training objective. Recent hybrid approaches augment CLIP's architecture with a multi-modal transformer \citep{albef,merlot}.

\paragraph{Zero-shot application of pre-trained models} Models pre-trained with the contrastive objective have exhibited strong zero-shot performance in image classification tasks \citep{clip,align}. \citet{vild} use CLIP can be to classify objects by computing scores for class labels with cropped proposals. Our IPS is different in that it isolates proposals by both cropping \emph{and blurring}. \citet{Shen2021HowMC} show that a simple zero-shot application of CLIP to visual question answering performs almost on par with random chance. \citet{cpt} describe a zero-shot method for ReC based on a pre-trained masked language model (MLM); we show that their zero-shot results and a version of their method adapted for models pre-trained to compute image-text scores (rather than MLM) are substantially worse than isolated proposal scoring and GradCAM. Concurrent with our work, \citet{NEURIPS2021_c3008b2c} also observe that CLIP has poor zero-shot accuracy when dealing with spatial relations.
\section{Conclusion}
\label{sec:conclusion}
We present ReCLIP, a zero-shot method for referring expression comprehension (ReC) that decomposes an expression into subqueries, uses CLIP to score isolated proposals against these subqueries, and combines the outputs with spatial heuristics. ReCLIP outperforms zero-shot ReC approaches from prior work and also performs well across visual domains: ReCLIP outperforms state-of-the-art supervised ReC models, trained on natural images, when evaluated on RefGTA. We also find that CLIP has low zero-shot spatial reasoning performance, suggesting the need for pre-training methods that account more for spatial reasoning.

\section{Ethical and Broader Impacts}
\label{sec:ethics}
Recent work has shown that pre-trained vision and language models suffer from biases such as gender bias \citep{ross-etal-2021-measuring,srinivasan2021worst}. \citet{agarwal2021evaluating} provide evidence that CLIP has racial and other biases, which makes sense since CLIP was trained on data collected from the web and not necessarily curated carefully. Therefore, we do not advise deploying our system directly in the real world immediately. Instead, practitioners interested in this system should first perform analysis to measure its biases based on previous work and attempt to mitigate them. We also note that our work relies heavily on a pre-trained model whose pre-training required a great deal of energy, which likely had negative environmental effects. That being said our zero-shot method does not require training a new model and in that sense could be more environmentally friendly than supervised ReC models (depending on the difference in the cost of inference).

\section{Acknowledgements}
We thank the Berkeley NLP group, Medhini Narasimhan, and the anonymous reviewers for helpful comments. We thank Michael Schmitz for help with AI2 infrastructure.
This work was supported in part by DoD, including DARPA’s LwLL (FA8750-19-1-0504), and/or SemaFor (HR00112020054) programs, and Berkeley Artificial Intelligence Research (BAIR) industrial alliance programs.
Sameer Singh was supported in part by the National Science Foundation grant \#IIS-1817183 and in part by the DARPA MCS program under Contract No. N660011924033 with the United States Office Of Naval Research.

\bibliography{anthology,custom-rebiber}
\bibliographystyle{acl_natbib}

\newpage
\appendix

\section{Visualization of Region-Scoring Methods}
\subsection{Colorful Prompt Tuning (CPT)}
\begin{figure}[t]
    \centering
    \includegraphics[width=0.4\textwidth]{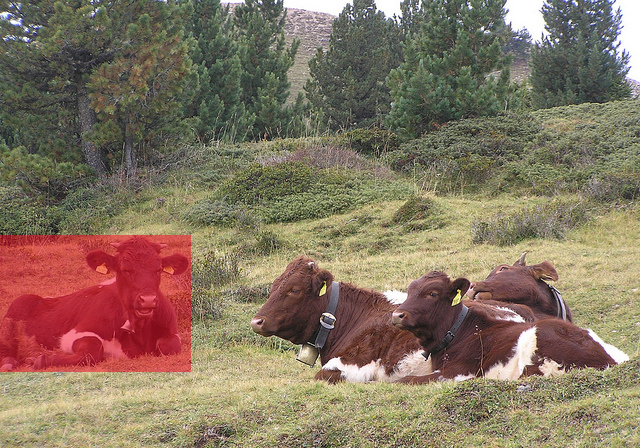}
    \caption{The visual representation of a proposal using CPT-adapted. The example is taken from the RefCOCOg validation set.}
    \label{fig:cpt_example}
\end{figure}

Figure~\ref{fig:cpt_example} shows an example of the visual representation of a proposal using CPT-adapted.

\subsection{Isolated Proposal Scoring (IPS)}
\label{app:ips_example}
\begin{figure}[t]
    \centering
    \includegraphics[width=0.4\textwidth]{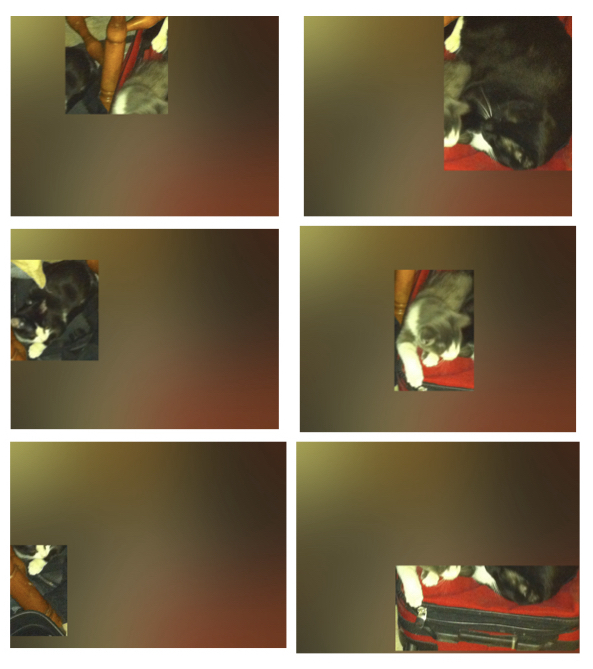}
    \caption{An example of isolating proposals by blurring the remainder of the image using $\sigma=100$}
    \label{fig:blur_example}
\end{figure}

Figure~\ref{fig:blur_example} shows the blurred versions of the proposals for an image using $\sigma=100$.

\section{Synthetic Spatial Reasoning Experiment}
\label{app:synthetic}
\begin{figure}[t]
    \centering
    \includegraphics[width=0.4\textwidth]{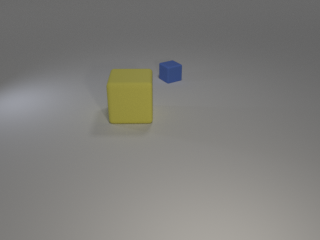}
    \caption{Example image for the synthetic text-pair tasks. For the spatial task, the text pair corresponding to this image is ``a yellow cube is in front of a blue cube.'' (correct) and ``a yellow cube is behind a blue cube.'' (incorrect). For the non-spatial (control) task, the text pair corresponding to this image is ``a blue cube and a yellow cube'' (correct) and ``a blue cube and a yellow sphere'' (incorrect).}
    \label{fig:text_pairs}
\end{figure}

Figure~\ref{fig:text_pairs} gives an example of the \emph{text-pairs} version of the synthetic tasks.

\label{app:cpt_example}

\begin{figure}[t]
    \centering
    \begin{subfigure}{0.5\textwidth}
        \centering
        \includegraphics[width=0.4\textwidth]{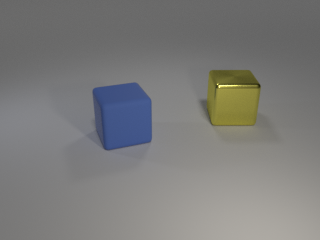} 
        \includegraphics[width=0.4\textwidth]{figures/CLEVR_new_000818.png}
        \caption{``a blue cube to the left of a yellow cube.''}
        \label{fig:image_pairs_subfig1}
    \end{subfigure}
    \begin{subfigure}{0.5\textwidth}
        \centering
        \includegraphics[width=0.4\textwidth]{figures/CLEVR_new_000637.png} 
        \includegraphics[width=0.4\textwidth]{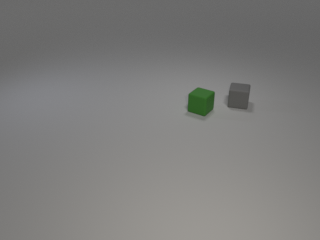}
        \caption{``a blue cube and a yellow cube''}
        \label{fig:image_pairs_subfig2}
    \end{subfigure}
    \caption{Examples of the image-pairs version of the spatial (\ref{fig:image_pairs_subfig1}) and non-spatial (\ref{fig:image_pairs_subfig2}) tasks. In each case, the left image is the correct one.}
    \label{fig:image_pairs}
\end{figure}
Figure~\ref{fig:image_pairs} gives an example of the \emph{image-pairs} version of the synthetic tasks.

\section{Semantic Formalism}

\subsection{Relation Semantics} \label{app:heuristics}
We use deterministic heuristics to compute the semantics of the following six relations: \emph{left}, \emph{right}, \emph{above}, \emph{below}, \emph{bigger}, and \emph{smaller}. On the other hand, we treat \emph{inside} as a random variable, and use heuristics to compute the value of its parameter.

For $R \in \{\emph{left}, \emph{right}, \emph{above}, \emph{below} \}$, we compute $R(i, j)$ by checking whether $R$ holds between the center point of box $i$ and box $j$. For example, if the center point of $i$ is to the left of the center point of box $j$, then $\emph{left}(i, j) = 1$.

We compute $\emph{bigger}(i, j)$ and $\emph{smaller}(i,j)$ simply by comparing the areas of boxes $i$ and $j$. For example, $\emph{bigger}(i, j)$ checks that the area of box $i$ is greater than the area of box $j$.

Finally, for $R = \emph{inside}$, we parameterize $r(i,j)$ as the ratio between the are of the intersection of boxes $i, j$ compared to the area of box $i$. Thus, unlike the other six deterministic rules, $\emph{inside}$ is modeled as a random variable.

\subsection{Relation Extraction} \label{app:extract}

We identify noun chunks in the dependency parse as predicates. We then extract relations by looking for dependency paths between the heads of noun chunks that contain the following keywords:
\begin{itemize}
    \item \emph{left}: ``left'', ``west''
    \item \emph{right}: ``right'', ``east''
    \item \emph{above}: ``above'', ``north'', ``top'', ``back'', ``behind''
    \item \emph{below}: ``below'', ``south'', ``under'', ``front''
    \item \emph{bigger}: ``bigger'', ``larger'', ``closer''
    \item \emph{smaller}: ``smaller'', ``tinier'', ``further''
    \item \emph{inside}: ``inside'', ``within'', ``contained''
\end{itemize}
We extract superlative relations by looking for dependency paths off the head of a noun chunk containing the following keywords:
\begin{itemize}
    \item \emph{left}: ``left'', ``west'', ``leftmost'', ``western''
    \item \emph{right}: ``right'', ``rightmost'', ``east'', ``eastern''
    \item \emph{above}: ``above'', ``north'', ``top''
    \item \emph{below}: ``below'', ``south'', ``underneath'', ``front''
    \item \emph{bigger}: ``bigger'', ``biggest'', ``larger'', ``largest'', ``closer'', ``closest''
    \item \emph{smaller}: ``smaller'', ``smallest'', ``tinier'',  ``tiniest'', ``further'', ``furthest''
\end{itemize}

\section{Description of ALBEF}
\label{app:albef_details}
The ALBEF model has an image-only transformer and a text-only transformer like CLIP but also has a multi-modal transformer that operates on the outputs of these two transformers. ALBEF is pre-trained with three losses: (1) an image-text contrastive (ITC) loss that works just like CLIP's and uses the outputs of the image-only and text-only transformers, (2) an image-text matching (ITM) loss--where the task is to decide whether a given image-text pair match--which uses the outputs of the multi-modal encoder, and (3) a masked language modeling loss which uses the outputs of the multi-modal encoder. We explore both the ITC and ITM scores in our experiments. ALBEF was pre-trained on roughly 15M image-caption pairs from conceptual captions \citep{conceptualcaptions}, SBU Captions \citep{sbucaptions}, COCO \citep{coco}, and Visual Genome \citep{visualgenome}.\footnote{As noted by the ALBEF authors, validation/test images of RefCOCO+ and RefCOCOg are included in the training set of COCO.}

\subsection{ALBEF Performance on Synthetic Spatial Reasoning Experiment}
\label{app:albef_relations}
Table~\ref{tab:albef_relations} shows the zero-shot accuracy of ALBEF ITM and ITC in the synthetic spatial reasoning experiment described in \S\ref{sec:clevr_exp}.

\begin{table}[t]
\centering
\captionsetup{font=footnotesize}
\resizebox{1.\linewidth}{!}
{
\begin{tabular}{lcccc}
\toprule
\multirow{2}[1]{*}{Model} & \multicolumn{1}{c}{Text-pair} & \multicolumn{1}{c}{Text-pair} & \multicolumn{1}{c}{Image-pair} & \multicolumn{1}{c}{Image-pair} \\
& \multicolumn{1}{c}{Spatial} & \multicolumn{1}{c}{Non-spatial} & \multicolumn{1}{c}{Spatial} & \multicolumn{1}{c}{Non-spatial} \\
\midrule
ALBEF ITM & 49.83 & 92.20 & 53.74 & 90.75 \\
ALBEF ITC & 49.83 & 85.42 & 51.54 & 72.25 \\
\bottomrule
\end{tabular}
}
\caption{Accuracy on CLEVR image-text matching task. ALBEF performs well on the non-spatial version of the task but poorly on the spatial version. Text-pair tasks have 295 instances each; image-pair tasks have 227 instances each.}
\label{tab:albef_relations}
\end{table}

\section{Implementation Details}
\label{app:implementation}
\subsection{Text prompt} For ALBEF, we pass the input expression directly to the model, whereas for CLIP, when using GradCAM and ReCLIP (with or without relations), we use the prefix ``a photo of'' following the authors' observations \citep{clip}. For CPT, the prompt is given in \S~\ref{sec:cpt}.
\subsection{Position embeddings} Both CLIP and ALBEF use fixed-size position embeddings, so either the input image must be resized to fit the dimensions of the embeddings or the size of the embeddings must be changed. For all models, we resize the image to match the model's visual input resolution. Resizing of images is done via bicubic interpolation. Figure~\ref{fig:gradcam_sensitivity} shows the how the performance of the GradCAM method varies between resizing images and resizing embeddings--for CLIP RN50x16, there is very little difference, while for CLIP ViT-B/32 image resizing makes a larger difference.

\subsection{GradCAM Layer}
For CLIP ViT-B/32, we use the last layer of the visual transformer for GradCAM. For CLIP RN50x16, we use output of layer 4 for GradCAM. For ALBEF ITM, we use the third layer of the multi-modal transformer for GradCAM (following \citet{albef}). For ALBEF ITC, we use the final layer of the visual transformer for GradCAM.

\subsection{Hyperparameter sensitivity}
\label{app:hyperparams}
\begin{figure}[t]
    \centering
    \includegraphics[width=0.49\textwidth]{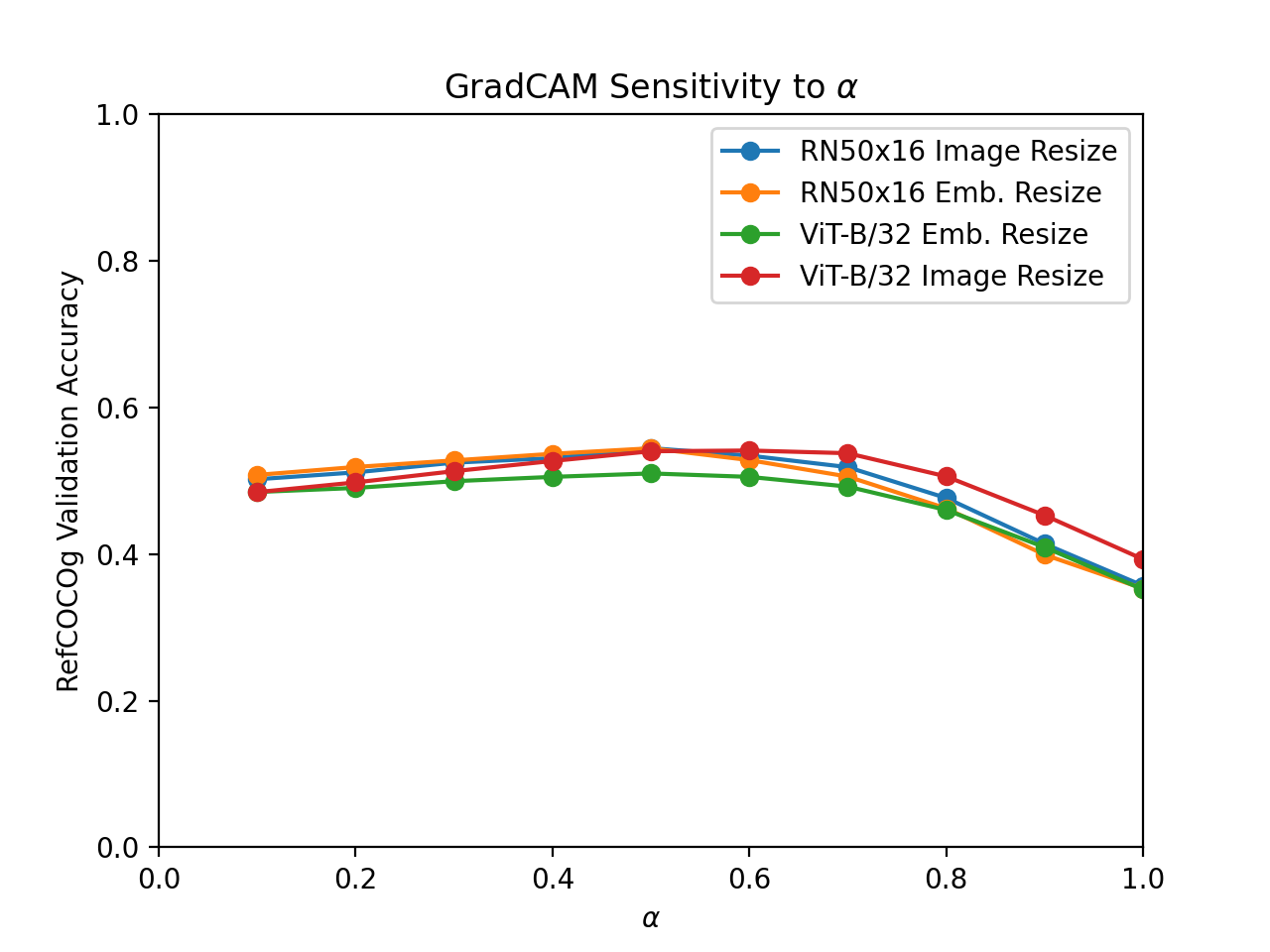}
    \caption{CLIP RN50x16 and ViT-B/32 Performance using GradCAM on RefCOCOg validation set comparing resizing of images with resizing of position embeddings, across 10 values of $\alpha$. These results use ground-truth proposals.}
    \label{fig:gradcam_sensitivity}
\end{figure}

\begin{figure}[t]
    \centering
    \includegraphics[width=0.49\textwidth]{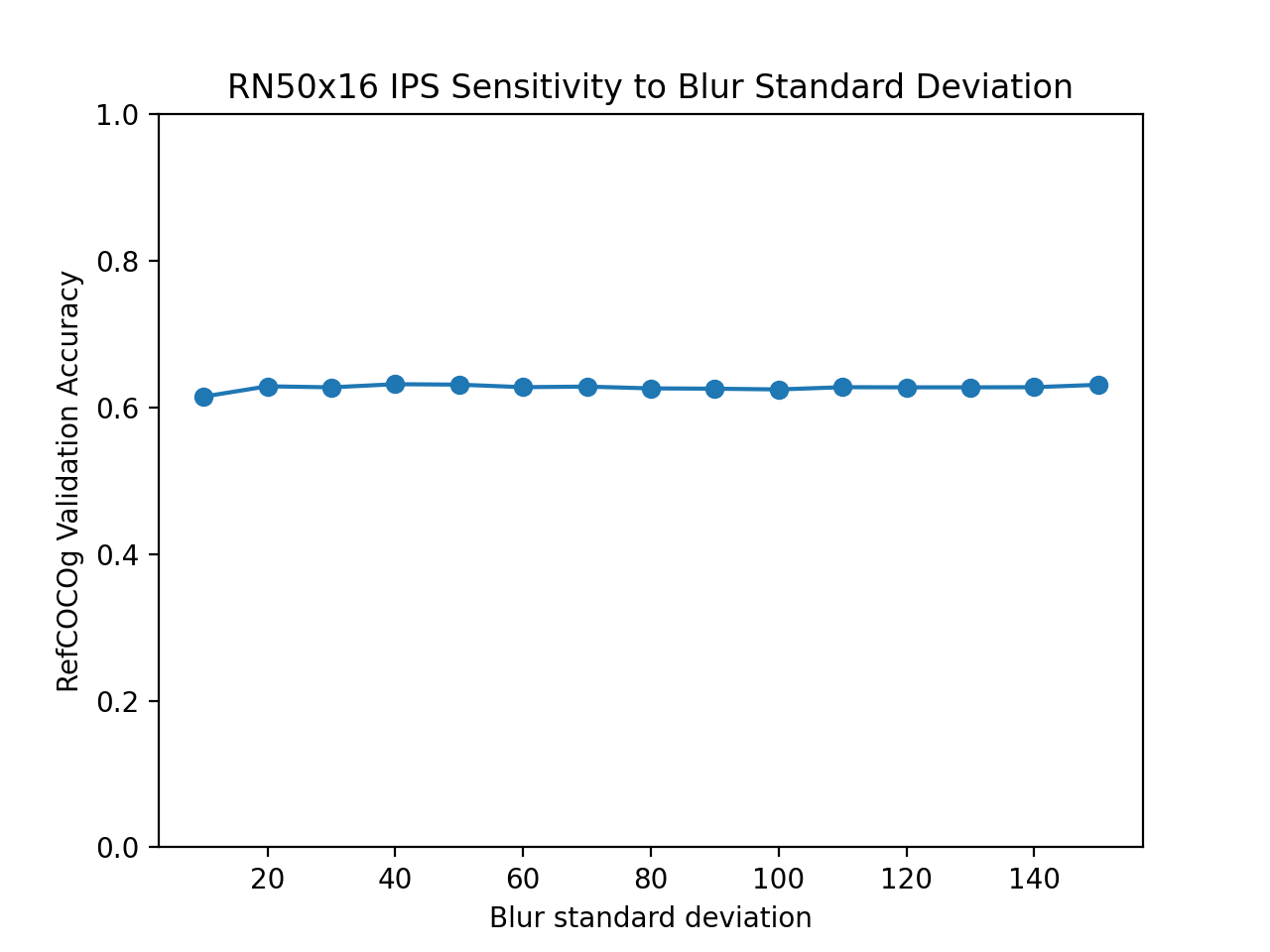}
    \caption{CLIP RN50x16 Performance using IPS on RefCOCOg validation set for different values of blur standard deviation $\sigma$. These results use ground-truth proposals.}
    \label{fig:blur_sensitivity}
\end{figure}

Figure~\ref{fig:gradcam_sensitivity} shows the sensitivity of the GradCAM method to $\alpha$ for the two CLIP models. We choose $\alpha=0.5$ for all models (including ALBEF), which results in the best accuracy for almost models. For ViT-B/32, $\alpha=0.6$ yields slightly higher accuracy by ($0.1\%$) on the RefCOCOg validation set. Figure~\ref{fig:blur_sensitivity} shows the sensitivity of the IPS method to the blur standard deviation $\sigma$ for the CLIP RN50x16 model. As shown, the method has little sensitivity to $\sigma$ above $\sigma=20$.

\subsection{Experimentation on validation set}
\label{app:experimentation}
As discussed by \citet{Perez2021TrueFL}, research on the zero-shot setting often uses labeled data for model selection. Aside from variants of IPS documented in our ablation study (\S\ref{sec:ablations}), we also experimented on the RefCOCOg validation set (and to a lesser extent on the RefCOCO+ validation set) with:
\begin{enumerate}
    \item Drawing a rectangle around the proposal and using an appropriate text prompt. Performance was somewhat similar to CPT performance.
    \item Ensembling the original text prompt with a text prompt having only the noun chunk of the expression containing the head word. This helped for IPS and is in a sense part of our rule-based relation-handling.
    \item Other techniques for handling superlatives. For instance, we tried to compute $\Pr[P_N(i) \wedge \bigwedge_{j \neq i} (\neg P_{N}(j) \vee (P_{N}(j) \wedge R(i, j)))]$. This performed worse than our chosen technique on the RefCOCOg validation set.
    \item Invoking the parser and relation-handling pipeline on all sentences rather than only those containing one of the relation/superlative keywords.
\end{enumerate}
We also selected the relation types and keywords based on these validation sets. Most of these preliminary experiments were performed using the area threshold mentioned in \S\ref{sec:coco_results}.

\subsection{Description of Computing Infrastructure}
We primarily used a machine with Quadro RTX 8000 GPUs, Google Cloud machines with V100 GPUs, and a machine with TITAN RTX and GeForce 2080s. These machines used Ubuntu as the operating system.

\subsection{Dataset Information}
\label{app:dataset_info}
All datasets that we use are focused on English. The COCO dataset can be downloaded from \url{https://cocodataset.org/#download}. The RefCOCO/g/+ datasets can be downloaded from \url{https://github.com/lichengunc/refer/tree/master/data}. The RefGTA dataset can be downloaded from \url{https://github.com/mikittt/easy-to-understand-REG/tree/master/pyutils/refer2}. The RefCOCOg validation set has 4896 instances, the RefCOCOg test set has 9602 instances, the RefCOCO+ validation set has 10758 instances, the RefCOCO+ TestA set has 5726 instances, the RefCOCO+ TestB set has 4889 instances, the RefCOCO validation set has 10834 instances, the RefCOCO TestA set has 5657 instances, the RefCOCO TestB set has 5095 instances, the RefGTA validation set has 17766 instances, and the RefGTA test set has 17646 instances.

\section{Qualitative Examples}
\label{app:qualitative_examples}
\begin{figure}[t]
    \centering
    \begin{subfigure}{0.24\textwidth}
        \includegraphics[width=\textwidth]{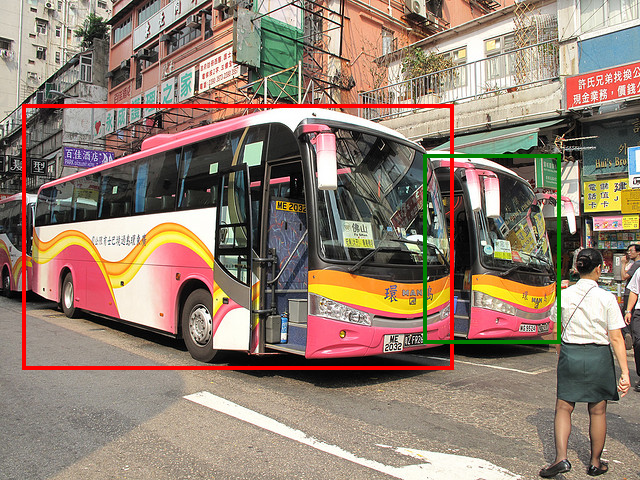}
        \caption{bus behind bus}
        \label{fig:bus_behind_bus}
    \end{subfigure}
    \begin{subfigure}{0.24\textwidth}
        \includegraphics[width=\textwidth]{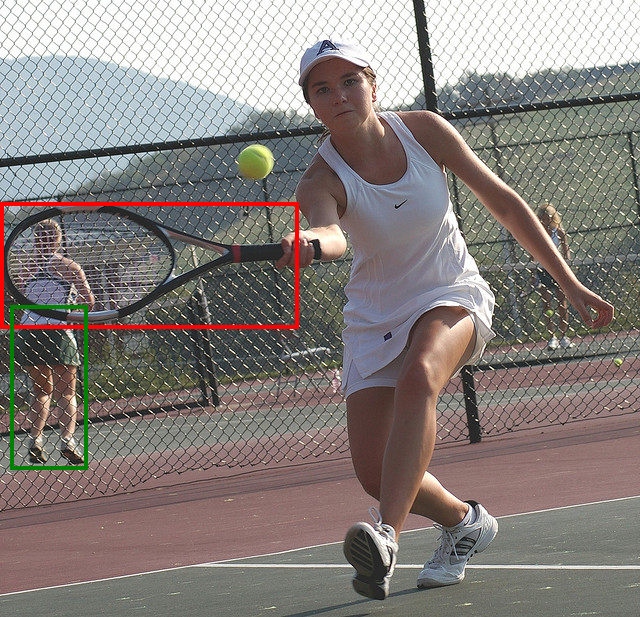}
        \caption{person behind the fence}
        \label{fig:person_behind_fence}
    \end{subfigure}
    \begin{subfigure}{0.24\textwidth}
        \includegraphics[width=\textwidth]{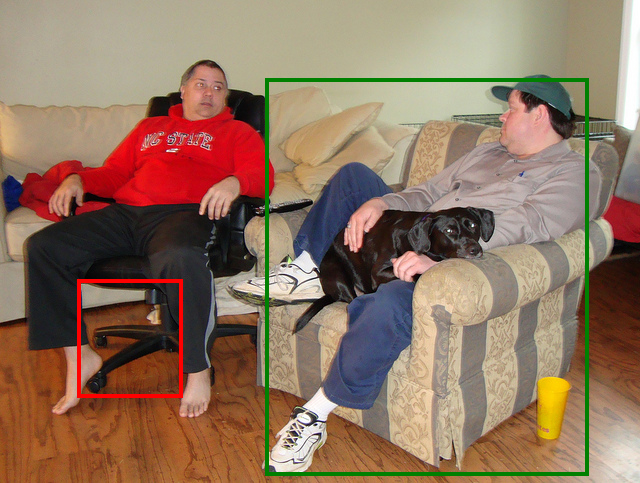}
        \caption{chair under dog}
        \label{fig:chair_under_dog}
    \end{subfigure}
    \begin{subfigure}{0.24\textwidth}
        \includegraphics[width=\textwidth]{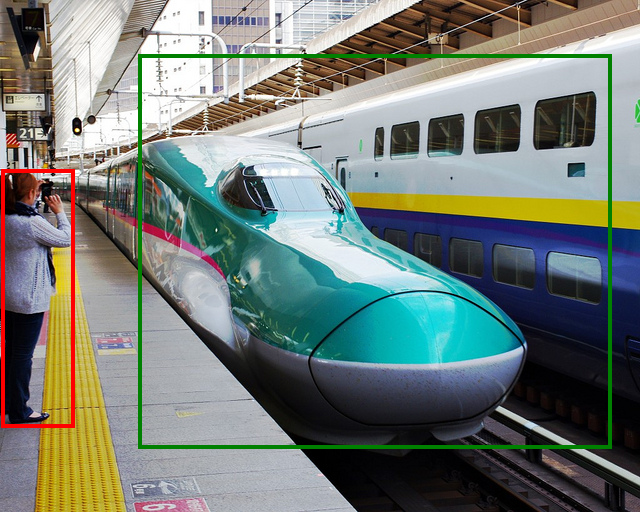}
        \caption{smallest train}
        \label{fig:smallest_train}
    \end{subfigure}
    \caption{Examples of execution errors causing ReCLIP to answer incorrectly on instances that it answers correctly when not using the relation-handling method. Parts \ref{fig:bus_behind_bus} and \ref{fig:person_behind_fence} show cases where the meaning of ``behind'' does not match our heuristic, which checks which proposal's $y$-coordinate is smaller. Part \ref{fig:chair_under_dog} shows an example where ``under'' means ``directly under.'' Part \ref{fig:smallest_train} shows an example in which due to the superlative ``smallest,'' the size of proposals appears to be weighted more heavily by our approach than scores CLIP assigns to the proposals based on the text.}
    \label{fig:error_examples}
\end{figure}

\begin{figure*}[t]
    \centering
    \begin{subfigure}{0.45\textwidth}
        \includegraphics[width=\textwidth]{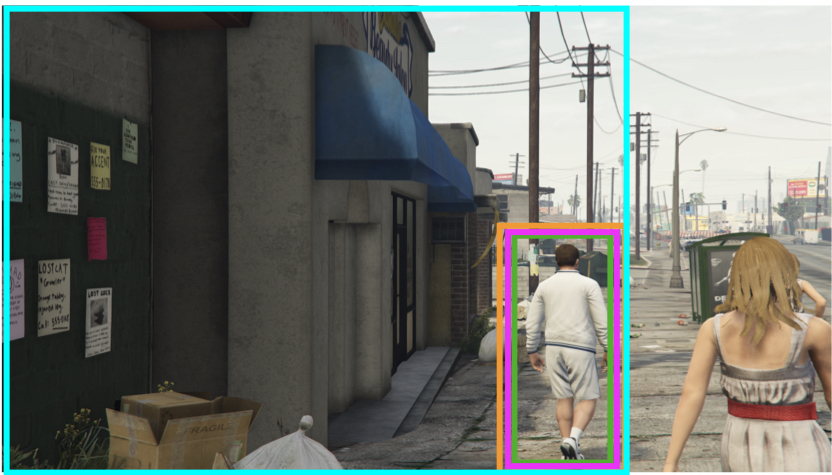}
        \caption{a man in white shorts and white jacket, walking down a sidewalk.}
    \end{subfigure}
    \begin{subfigure}{0.45\textwidth}
        \includegraphics[width=\textwidth]{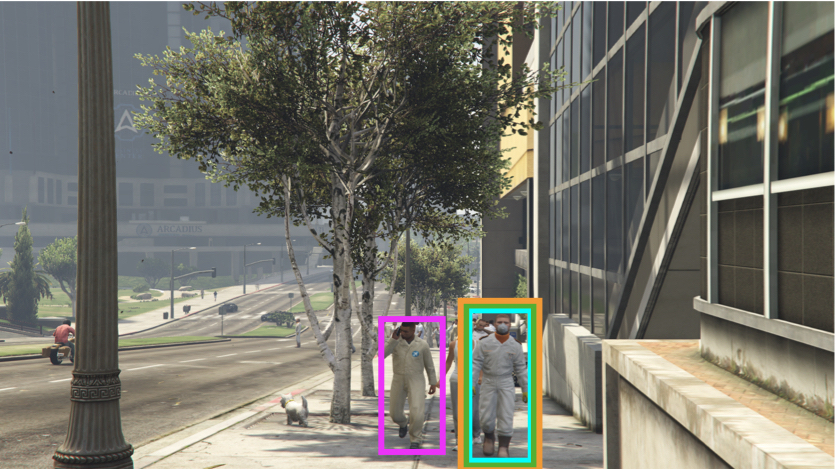}
        \caption{a man in white jumpsuit with face mask walking.}
    \end{subfigure}
    \begin{subfigure}{0.45\textwidth}
        \includegraphics[width=\textwidth]{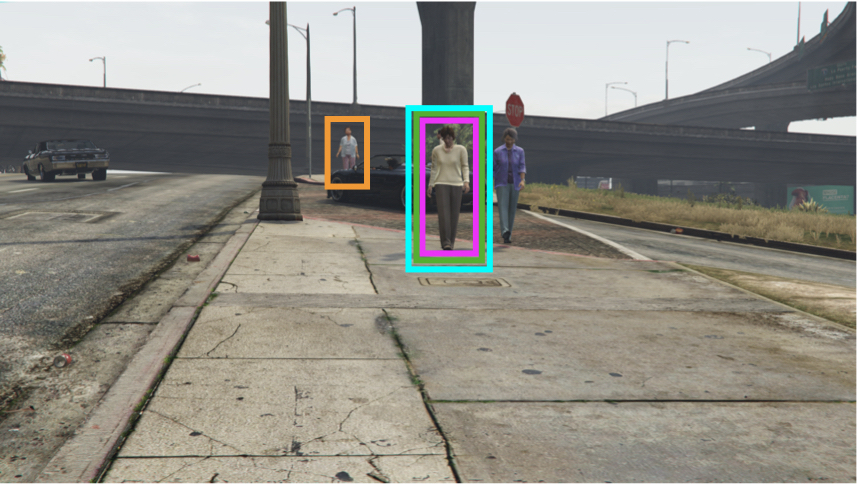}
        \caption{an african american woman with light colored sweater, brown pants walking down sidewalk near another woman.}
    \end{subfigure}
    \begin{subfigure}{0.45\textwidth}
        \includegraphics[width=\textwidth]{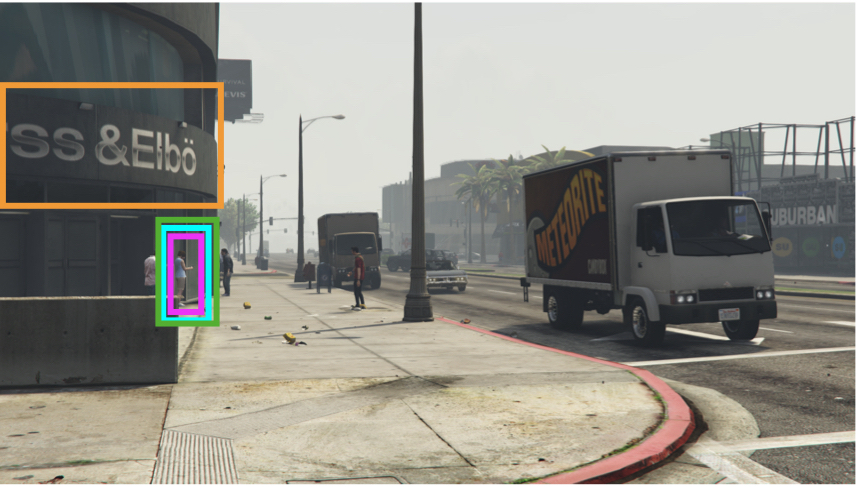}
        \caption{woman in blue shirt in doorway.}
    \end{subfigure}
    \begin{subfigure}{0.45\textwidth}
        \includegraphics[width=\textwidth]{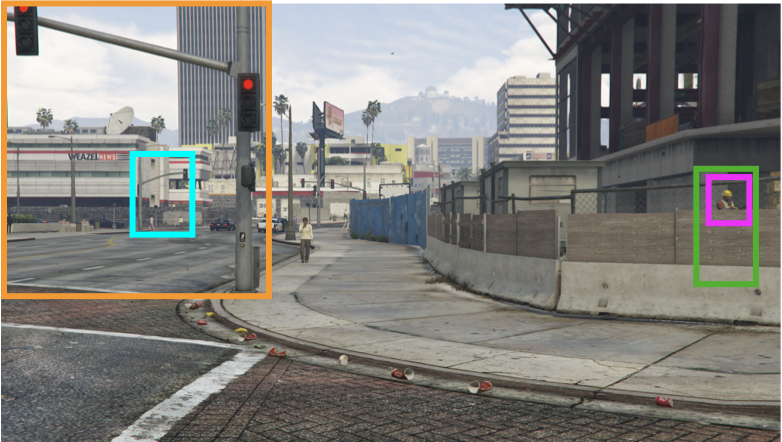}
        \caption{a man with yellow helmet behind the fence.}
    \end{subfigure}
    \begin{subfigure}{0.45\textwidth}
        \includegraphics[width=\textwidth]{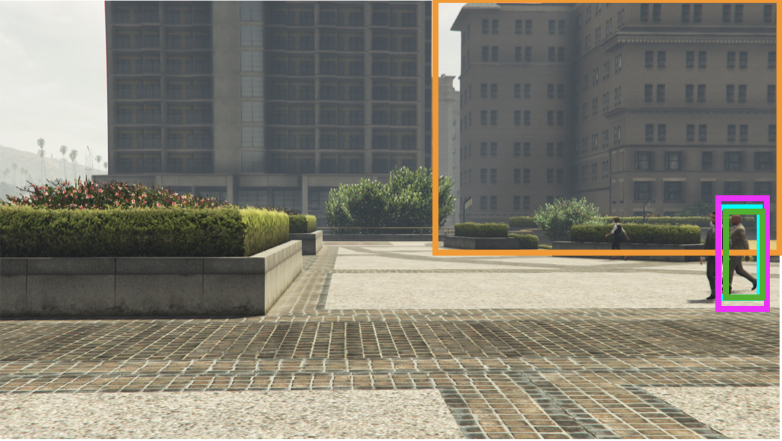}
        \caption{a bald black man is walking wearing a tan suit.}
    \end{subfigure}
    \begin{subfigure}{0.45\textwidth}
        \includegraphics[width=\textwidth]{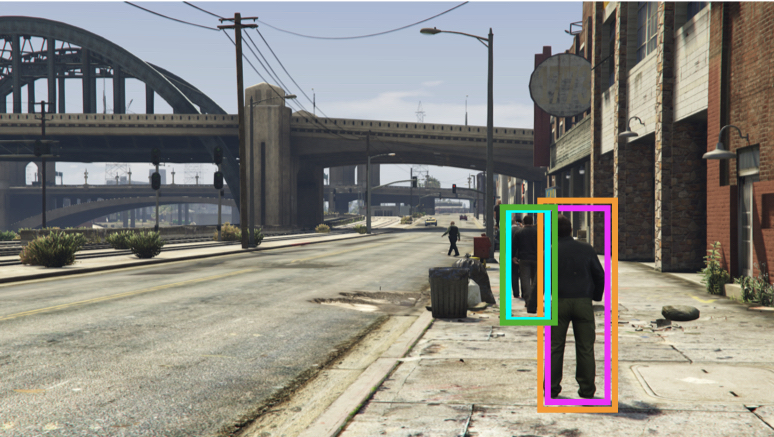}
        \caption{a man in all black walking in front of another man.}
    \end{subfigure}
    \begin{subfigure}{0.45\textwidth}
        \includegraphics[width=\textwidth]{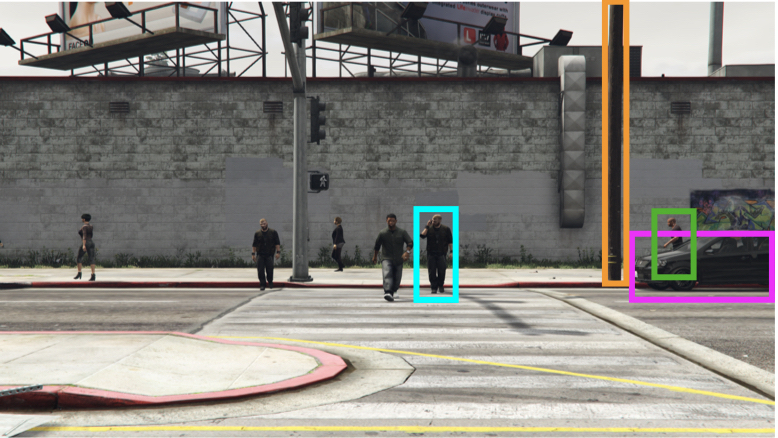}
        \caption{a man wearing a short-sleeved black top walks by a black car.}
    \end{subfigure}
    \begin{subfigure}{0.45\textwidth}
        \includegraphics[width=\textwidth]{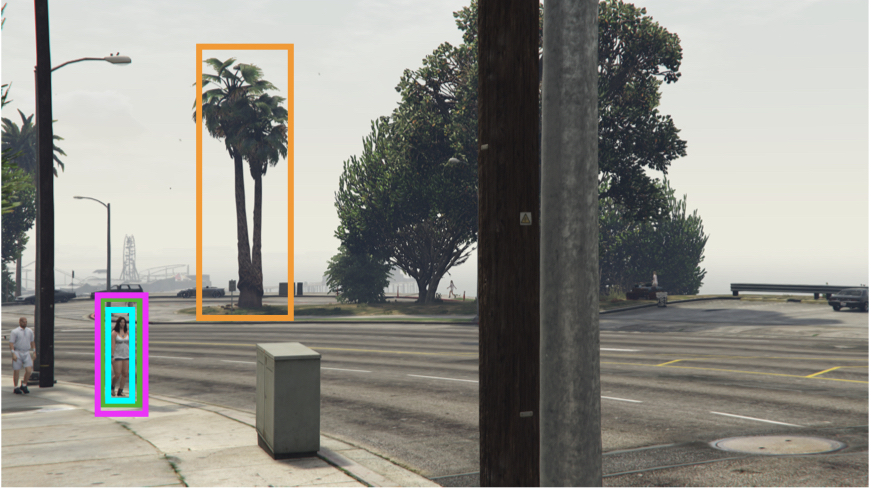}
        \caption{a woman in a white top.}
    \end{subfigure}
    \begin{subfigure}{0.45\textwidth}
        \includegraphics[width=\textwidth]{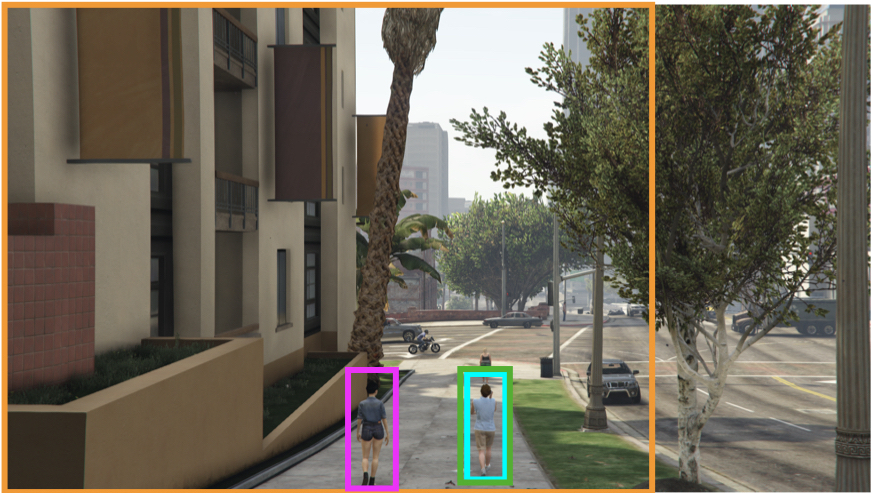}
        \caption{a man in a blue polo and brown shorts talking on a cell phone.}
    \end{subfigure}
    \caption{Qualitative examples randomly sampled from the RefGTA validation set. Ground-truth referents are in \textcolor{green}{green}, MDETR (pre-trained) predictions are in \textcolor{magenta}{magenta}, UNITER (trained on RefCOCO+) predictions are in \textcolor{orange}{orange}, and ReCLIP predictions are in \textcolor{cyan}{cyan}. The subcaptions are the corresponding referring expressions. For UNITER and ReCLIP, this represents the setting in which we consider all proposals from UniDet.}
    \label{fig:gta_examples}
\end{figure*}

Figure~\ref{fig:gta_examples} shows qualitative examples for the RefGTA validation set. Figure~\ref{fig:error_examples} shows examples of the execution errors mentioned in the error analysis in Section~\ref{sec:ablations}.

\section{Additional Experiment Results}
\label{app:results}
This section presents the full results on the RefCOCOg/RefCOCO+/RefCOCO datasets, including results without ensembling using CLIP RN50x16 and ViT-B/32 models and results using ground-truth proposals. Table~\ref{tab:coco-full-results} shows full results on the RefCOCOg and RefCOCO+ datasets. Table~\ref{tab:refcoco-full-results} shows full results on the RefCOCO dataset.
\begin{table*}[t]
\small
\centering
\captionsetup{font=footnotesize}
\resizebox{1.0\textwidth}{!}{
\begin{tabular}{lcccc|ccccccc}
\toprule
\multirow{2}[3]{*}{{\bf Model}} & \multicolumn{4}{c}{RefCOCOg} & \multicolumn{6}{c}{RefCOCO+} \\
& \multicolumn{1}{c}{\textbf{$Val_g$}} & \multicolumn{1}{c}{\textbf{$Val_d$}} & \multicolumn{1}{c}{\textbf{$Test_g$}} & \multicolumn{1}{c}{\textbf{$Test_d$}} & \multicolumn{1}{c}{\textbf{$Val_g$}} & \multicolumn{1}{c}{\textbf{$Val_d$}} & \multicolumn{1}{c}{\textbf{$TestA_g$}} & \multicolumn{1}{c}{\textbf{$TestA_d$}} & \multicolumn{1}{c}{\textbf{$TestB_g$}} & \multicolumn{1}{c}{\textbf{$TestB_d$}} \\
\midrule
Random & 20.18 & 18.117 & 20.34 & 19.10 & 16.73 & 16.29 & 12.57 & 13.57 & 22.13 & 19.60 \\
\midrule
UNITER-L (supervised; \citet{uniter}) & 87.85 & 74.86 & 87.73 & 75.77 & 84.25 & 75.90 & 86.34 & 81.45 & 79.75 & 75.77 & \\
MDETR (supervised; \citet{mdetr}) & -- & 83.35 & -- & 81.64 & -- & 81.13 & -- & 85.52 & -- & 72.96 & \\
\midrule
Weakly supervised (non-pretrained; \citet{Sun2021DiscriminativeTM}) & -- & -- & -- & -- & 39.18 & 38.91 & 40.01 & 39.91 & 38.08 & 37.09 & \\
\midrule
CPT-Blk w/ VinVL \citep{cpt} & -- & 32.1 & -- & 32.3 & -- & 25.4 & -- & 25.0 & -- & 27.0 & \\
CPT-Seg w/ VinVL \citep{cpt} & -- & 36.7 & -- & 36.5 & -- & 31.9 & -- & 35.2 & -- & 28.8 & \\
\midrule
\textbf{CLIP RN50x16} & & & & & & & & & & & \\
CPT-adapted & 27.74 & 25.04 & 28.81 & 25.92 & 24.48 & 22.09 & 20.22 & 19.54 & 27.80 & 25.57 & \\
GradCAM & 54.51 & 48.35 & 53.71 & 47.50 & \textbf{48.29} & \textbf{44.53} & \textbf{52.86} & \textbf{52.78} & 41.13 & 35.67 & \\
ReCLIP w/o relations & 62.50 & 55.88 & 62.03 & 54.33 & 47.12 & 44.15 & 46.47 & 45.97 & 49.62 & 41.79 & \\
ReCLIP & \textbf{64.79} & \textbf{57.66} & \textbf{64.39} & \textbf{56.37} & \textbf{47.92} & \textbf{44.53} & 46.38 & 45.88 & \textbf{50.89} & \textbf{42.87} & \\
\midrule
\textbf{CLIP ViT-B/32} & & & & & & & & & & & \\
CPT-adapted & 24.16 & 21.77 & 24.70 & 22.78 & 25.07 & 23.46 & 22.28 & 21.73 & 28.68 & 26.32 & \\
GradCAM & 54.00 & 49.51 & 54.01 & 48.53 & 48.00 & 44.64 & \textbf{52.13} & \textbf{50.73} & 43.85 & 39.01 & \\
ReCLIP w/o relations & 62.38 & 55.35 & 61.76 & 54.33 & 48.53 & 44.96 & 50.16 & 48.24 & 47.29 & 41.71 & \\
ReCLIP w/o relations & \textbf{65.48} & \textbf{56.96} & \textbf{64.38} & \textbf{56.15} & \textbf{49.20} & \textbf{45.34} & 50.23 & 48.45 & \textbf{48.58} & \textbf{42.71} & \\
\midrule
\textbf{CLIP Ensemble} & & & & & & & & & & & \\
CPT-adapted & 25.96 & 22.32 & 25.87 & 23.65 & 25.44 & 23.85 & 22.00 & 21.55 & 28.74 & 25.92 & \\
GradCAM & 56.82 & 50.86 & 56.15 & 49.70 & 51.10 & 47.83 & \textbf{57.79} & \textbf{56.92} & 43.24 & 37.70 & \\
ReCLIP w/o relations & 65.32 & 57.70 & 65.59 & 57.19 & 51.54 & 47.43 & 51.80 & 50.02 & 50.85 & 43.85 & \\
ReCLIP & \textbf{68.08} & \textbf{59.33} & \textbf{67.05} & \textbf{59.01} & \textbf{52.12} & \textbf{47.87} & 51.61 & 50.10 & \textbf{52.03} & \textbf{45.10} & \\
\bottomrule
\end{tabular}
}
\caption{Accuracy on the RefCOCOg and RefCOCO+ datasets. ReCLIP outperforms other zero-shot methods on RefCOCOg. On RefCOCO+, ReCLIP is roughly on par with GradCAM but has lower variance between TestA and TestB, which correspond to different kinds of objects. Subscript \emph{g} indicates ground-truth proposals are used, and \emph{d} indicates detected proposals are used. Best zero-shot results for each model and each column are in \textbf{bold}. See Table~\ref{tab:coco-results} for results using object size prior.}
\label{tab:coco-full-results}
\end{table*}

\begin{table*}[t]
\small
\centering
\captionsetup{font=footnotesize}
\resizebox{1.0\textwidth}{!}{
\begin{tabular}{lcccccc}
\toprule
\multirow{2}[3]{*}{{\bf Model}} & \multicolumn{6}{c}{RefCOCO} \\
& \multicolumn{1}{c}{\textbf{$Val_g$}} & \multicolumn{1}{c}{\textbf{$Val_d$}} & \multicolumn{1}{c}{\textbf{$TestA_g$}} & \multicolumn{1}{c}{\textbf{$TestA_d$}} & \multicolumn{1}{c}{\textbf{$TestB_g$}} & \multicolumn{1}{c}{\textbf{$TestB_d$}} \\
\midrule
Random & 16.37 & 15.73 & 12.45 & 13.51 & 21.32 & 19.20 \\
\midrule
UNITER-L (supervised; \citet{uniter}) & 91.84 & 81.41 & 92.65 & 87.04 & 91.19 & 74.17 \\
MDETR (supervised; \citet{mdetr}) & -- & 87.51 & -- & 90.40 & -- & 82.67 \\
\midrule
Weakly supervised (non-pretrained; \citet{Sun2021DiscriminativeTM}) & 39.21 & 38.35 & 41.14 & 39.51 & 37.72 & 37.01 \\
\midrule
CPT-Blk w/ VinVL \citep{cpt} & -- & 26.9 & -- & 27.5 & -- & 27.4  \\
CPT-Seg w/ VinVL \citep{cpt} & -- & 32.2 & -- & 36.1 & -- & 30.3  \\
\midrule
\textbf{CLIP RN50x16} & & & & & & \\
CPT-adapted & 23.31 & 21.48 & 19.25 & 18.56 & 28.36 & 25.28 \\
GradCAM & 44.00 & 40.49 & \textbf{47.41} & \textbf{46.51} & 38.17 & 33.66 \\
ReCLIP w/o relations & 40.62 & 37.61 & 39.08 & 38.39 & 43.55 & 37.17 \\
ReCLIP & \textbf{45.94} & \textbf{41.53} & 41.24 & 40.78 & \textbf{52.64} & \textbf{45.55} \\
\midrule
\textbf{CLIP ViT-B/32} & & & & & & \\
CPT-adapted & 25.12 & 23.79 & 23.39 & 22.87 & 28.42 & 26.03 \\
GradCAM & 45.41 & 42.29 & \textbf{50.13} & \textbf{49.04} & 41.47 & 36.68  \\
ReCLIP w/o relations & 44.37 & 40.58 & 45.09 & 43.98 & 43.42 & 37.63 \\
ReCLIP & \textbf{49.69} & \textbf{45.77} & 48.08 & 46.99 & \textbf{52.50} & \textbf{45.24} \\
\midrule
\textbf{CLIP Ensemble} & & & & & & \\
CPT-adapted & 24.79 & 23.16 & 21.62 & 21.44 & 28.89 & 26.95 \\
GradCAM & 46.68 & 42.85 & \textbf{51.99} & \textbf{51.07} & 40.10 & 35.21 \\
ReCLIP w/o relations & 45.66 & 41.97 & 45.13 & 43.42 & 45.40 & 39.02 \\
ReCLIP & \textbf{50.51} & \textbf{45.78} & 47.11 & 46.10 & \textbf{54.94} & \textbf{47.07} \\
\bottomrule
\end{tabular}
}
\caption{Accuracy on the RefCOCO dataset. Subscript \emph{g} indicates ground-truth proposals are used, and \emph{d} indicates detected proposals are used. Best zero-shot results for each model and each column are in \textbf{bold}. See Table~\ref{tab:coco-results} for results using object size prior.}
\label{tab:refcoco-full-results}
\end{table*}



\end{document}